%% file: iclr2027_conference.tex
\PassOptionsToPackage{table}{xcolor}

\documentclass{article}
\usepackage{iclr2027_conference,times}

\input{math_commands.tex}

\usepackage{amsmath}
\usepackage{amssymb}
\usepackage{amsfonts}
\usepackage{amsthm}
\usepackage{mathtools}
\usepackage{bm}
\usepackage{bbm}
\usepackage{nicefrac}

\usepackage{algorithm}
\usepackage{algpseudocode}

\usepackage{array}
\usepackage{booktabs}
\usepackage{multirow}
\usepackage{threeparttable}
\usepackage{adjustbox}
\usepackage{xcolor}
\usepackage{colortbl}

\usepackage{graphicx}
\usepackage{wrapfig}
\usepackage{float}

\usepackage{tikz}
\usepackage{pgfplots}
\pgfplotsset{compat=1.18}

\usepgfplotslibrary{fillbetween}
\usepgflibrary{patterns}

\usetikzlibrary{
    arrows.meta,
    positioning,
    calc,
    shapes.geometric,
    decorations.text,
    patterns,
    patterns.meta,
    fit,
    backgrounds,
    chains,
    shadows,
    math,
    circuits.ee.IEC,
    decorations.pathmorphing,
    decorations.pathreplacing,
    decorations.shapes
}

\usepackage[labelfont=bf,font=small]{caption}
\usepackage{subcaption}
\usepackage{listings}

\usepackage{textcomp}
\usepackage{pifont}
\usepackage{scalefnt}
\usepackage{tcolorbox}
\tcbuselibrary{skins}

\usepackage{hyperref}
\usepackage{url}

\definecolor{codebg}{RGB}{245,245,245}
\definecolor{keywordcolor}{RGB}{0,0,153}
\definecolor{commentcolor}{RGB}{34,139,34}
\definecolor{stringcolor}{RGB}{163,21,21}
\definecolor{numbercolor}{RGB}{128,128,128}

\title{PHBA: Prefix-State Hybrid Block Attention}

\author{
\textbf{
Ruijie Li$^{1}$,
Jiaxi Hu$^{1,*}$,
Shiyu Wang$^{2,*}$,
Yuxuan Liang$^{1,*}$
} \\
\normalfont
$^{1}$The Hong Kong University of Science and Technology (Guangzhou) \\
$^{2}$Independent researcher \\
$^{*}$Corresponding author
}

\iclrfinalcopy

\begin{document}

\maketitle

\begin{abstract}
Hybrid architectures combining linear sequence models with softmax attention provide an effective balance between efficient long-context modeling and precise token retrieval. Existing designs such as Native Hybrid Attention (NHA) combine compressed long-term states with sliding-window attention, but their exact attention is restricted to a fixed local window. In this work, we introduce Prefix-State Hybrid Block Attention (PHBA), which replaces local sliding-window attention with top-$k$ block-sparse retrieval and couples each retrieved block with a compact prefix state summarizing its preceding context. The prefix states are constructed by a gated linear recurrence at block boundaries and retrieved together with the corresponding token blocks, allowing the model to combine precise long-range evidence with compressed historical context within a unified layer. We further develop a hardware-aware Triton implementation that streams routed token blocks and prefix states without materializing large intermediate tensors. Experiments show that PHBA improves long-context and retrieval performance over strong linear and hybrid baselines while retaining efficient training and inference.
\end{abstract}

\section{Introduction}

Long-context language modeling requires both efficient sequence processing and
reliable retrieval of information from distant positions. Standard
self-attention~\citep{vaswani2017attention,dao2022flashattention} provides precise token-level access to the entire history, but
its quadratic complexity makes it increasingly expensive as the context length
grows.
Linear and recurrent sequence models~\citep{
katharopoulos2020transformers,
choromanski2020rethinking,
gu2021efficiently,
smith2022simplified,
sun2023retentive,
gu2023mamba,
dao2024transformers,
yang2023gated,
arora2024simple,
hu2025comba,
qin2024hgrn2,
zhang2024gated,
team2025kimi,
hatamizadeh2026gated} offer a more efficient alternative by
compressing historical information into fixed-size states
,
but this compression can limit precise in-context retrieval.

A practical solution is to combine recurrent sequence modeling with sparse or
local softmax attention. Recent hybrid architectures mix efficient recurrent
computation with token-level attention to improve the trade-off between
long-range efficiency and recall
~\citep{arora2024simple,de2024griffin,du2026native}.
For example, Native Hybrid Attention (NHA)~\citep{du2026native}
combines a compressed recurrent state with sliding-window attention (SWA),
allowing each layer to access both summarized long-term memory and precise
recent tokens. However, the exact-attention component in SWA is inherently
local: key--value pairs outside the fixed window are accessible only through
the compressed recurrent state. As a result, the exact-attention budget cannot
be allocated according to the content relevance between each query and the
global history. This leaves a central challenge for long-context hybrid models:

\begin{tcolorbox}[
    enhanced,
    width=0.95\linewidth,
    center,
    colback=white,
    colframe=gray!60!black,
    boxrule=1pt,
    arc=2mm,
    drop fuzzy shadow
]
\centering\itshape
Can sparse attention retrieve the globally relevant key--value content
for each query while preserving its preceding context?
\end{tcolorbox}

Sparse attention has long been explored as a way to extend token-level
interaction beyond dense quadratic attention
~\citep{
child2019generating,
kitaev2020reformer,
beltagy2020longformer,
zaheer2020big,
roy2021efficient}.
A natural way to extend exact attention beyond a fixed local window is
content-dependent block retrieval. Top-$K$ block-sparse attention allocates the
sparse computation to relevant blocks from distant history rather than only to
recent tokens. In particular, Mixture of Block Attention
(MoBA)~\citep{lu2026moba} uses query-dependent block-level routing to identify
a small number of historical blocks for each query.

Based on this observation, we introduce Prefix-State Hybrid Block Attention
(PHBA). PHBA replaces the fixed sliding-window component of conventional
recurrent--attention hybrids with content-dependent block-sparse retrieval.
In addition, PHBA associates each block with a compact prefix state constructed
by a gated slot recurrence~\citep{yang2023gated,zhang2024gated} at the
corresponding block boundary. When a query selects a historical block, it
accesses both the exact tokens within that block and its aligned prefix state,
which summarizes the causal history preceding the block. PHBA therefore couples
high-resolution retrieved content with compressed context immediately preceding
that content.

The key design is to align recurrent memory with the same block structure used
for sparse routing. Rather than relying on a single compressed history together
with fixed local attention, PHBA exposes block-aligned prefix memories that can
be selected together with distant exact-token blocks. This allows the model to
recover broader historical context around retrieved content without maintaining
a separate recurrent state for every token.

Efficiently implementing this computation requires processing dynamically
routed token blocks together with their associated prefix states without
materializing packed sparse payloads. We therefore develop a hardware-aware
Triton implementation that directly streams selected token and state blocks
from their original memory layouts and uses compact routing metadata for
efficient sparse execution. Combined with chunkwise prefix-state construction
and online-softmax accumulation, the implementation supports efficient
long-context training and inference. Experiments demonstrate that PHBA consistently improves retrieval and
long-context modeling over strong recurrent, sparse-attention, and hybrid
baselines while maintaining favorable efficiency at long sequence lengths.

\section{Preliminary}
\label{sec:preliminary}

PHBA builds upon two components: block-sparse attention for
selective token retrieval and gated slot memory for recurrent
sequence modeling. This section briefly reviews these two components
and introduces the notation used in PHBA.

\subsection{Block-Sparse Attention}
\label{sec:prelim-block-sparse}

Sparse attention reduces the cost of standard self-attention by allowing each
query to attend to only a subset of historical tokens. Block-sparse attention
groups the sequence into contiguous blocks and performs selection at the block
level.

Let the sequence be partitioned into blocks
$\{\mathcal{B}_1,\ldots,\mathcal{B}_B\}$, and let
$\mathcal{R}_t$ denote the set of blocks selected for query $\bm q_t$.
The query attends to the tokens contained in the selected blocks:
\begin{equation}
\bm{o}^{\mathrm{blk}}_t
=
\operatorname{softmax}
\left(
\frac{
\bm q_t
\bm K_{\mathcal{R}_t}^{\top}
}{
\sqrt d
}
\right)
\bm V_{\mathcal{R}_t},
\end{equation}
where $\bm K_{\mathcal{R}_t}$ and $\bm V_{\mathcal{R}_t}$ contain the
key and value vectors from the selected blocks.

Content-dependent block routing, as used in Mixture of Block Attention (MoBA),
selects a small number of relevant blocks for each query. PHBA adopts this
block-routed sparse-attention setting; the routing procedure is described in
Sec.~\ref{sec:method}.

\subsection{Gated Slot Memory}
\label{sec:prelim-gated-memory}

Gated Slot Attention (GSA)~\citep{zhang2024gated} represents historical
information using a fixed number of key and value memory slots.
Let
$\widetilde{\bm K}_t\in\mathbb{R}^{M\times d}$ and
$\widetilde{\bm V}_t\in\mathbb{R}^{M\times d_v}$ denote the key and value
memories, and let $\bm A_t\in(0,1)^M$ denote a slot-wise retention gate.
Using the coupled write gate $\bm I_t=\bm 1-\bm A_t$, the memories are
updated as
\begin{equation}
\begin{aligned}
\widetilde{\bm K}_t
&=
\operatorname{Diag}(\bm A_t)\widetilde{\bm K}_{t-1}
+
\bm I_t\bm k_t^\top,\\
\widetilde{\bm V}_t
&=
\operatorname{Diag}(\bm A_t)\widetilde{\bm V}_{t-1}
+
\bm I_t\bm v_t^\top .
\end{aligned}
\end{equation}
The shared slot-wise gate keeps the key and value memories temporally
aligned. This recurrence is a specialization of gated linear attention
(GLA)~\citep{yang2023gated} and admits the same hardware-efficient
chunkwise-parallel formulation. PHBA adopts this gated slot-memory construction to form block-aligned
prefix states, but uses them differently from GSA: the resulting key/value
slots are stored only at block boundaries and later participate in joint
softmax attention together with routed exact-token blocks.
% \subsection{Gated Linear Attention}
% \label{sec:prelim-gla}

% Linear attention can be expressed as a recurrent key--value memory.
% Given query, key, and value vectors
% $\bm q_t,\bm k_t,\bm v_t$, a linear recurrence maintains a matrix state
% $\bm S_t$ as
% \begin{equation}
% \bm S_t
% =
% \bm S_{t-1}
% +
% \bm v_t\bm k_t^\top,
% \qquad
% \bm o_t
% =
% \bm S_t\bm q_t .
% \end{equation}

% Gated Linear Attention (GLA) introduces a data-dependent decay to control
% the recurrent state:
% \begin{equation}
% \bm S_t
% =
% \bm G_t \odot \bm S_{t-1}
% +
% \bm v_t\bm k_t^\top,
% \qquad
% \bm o_t
% =
% \bm S_t\bm q_t ,
% \end{equation}
% where $\bm G_t$ is an input-dependent forget gate.
% With an appropriate factorized parameterization of $\bm G_t$, the recurrence
% admits hardware-efficient chunkwise-parallel training.

\section{Prefix-State Hybrid Block Attention}
\label{sec:method}

\begin{figure*}[t]
    \centering
    \includegraphics[width=\textwidth]{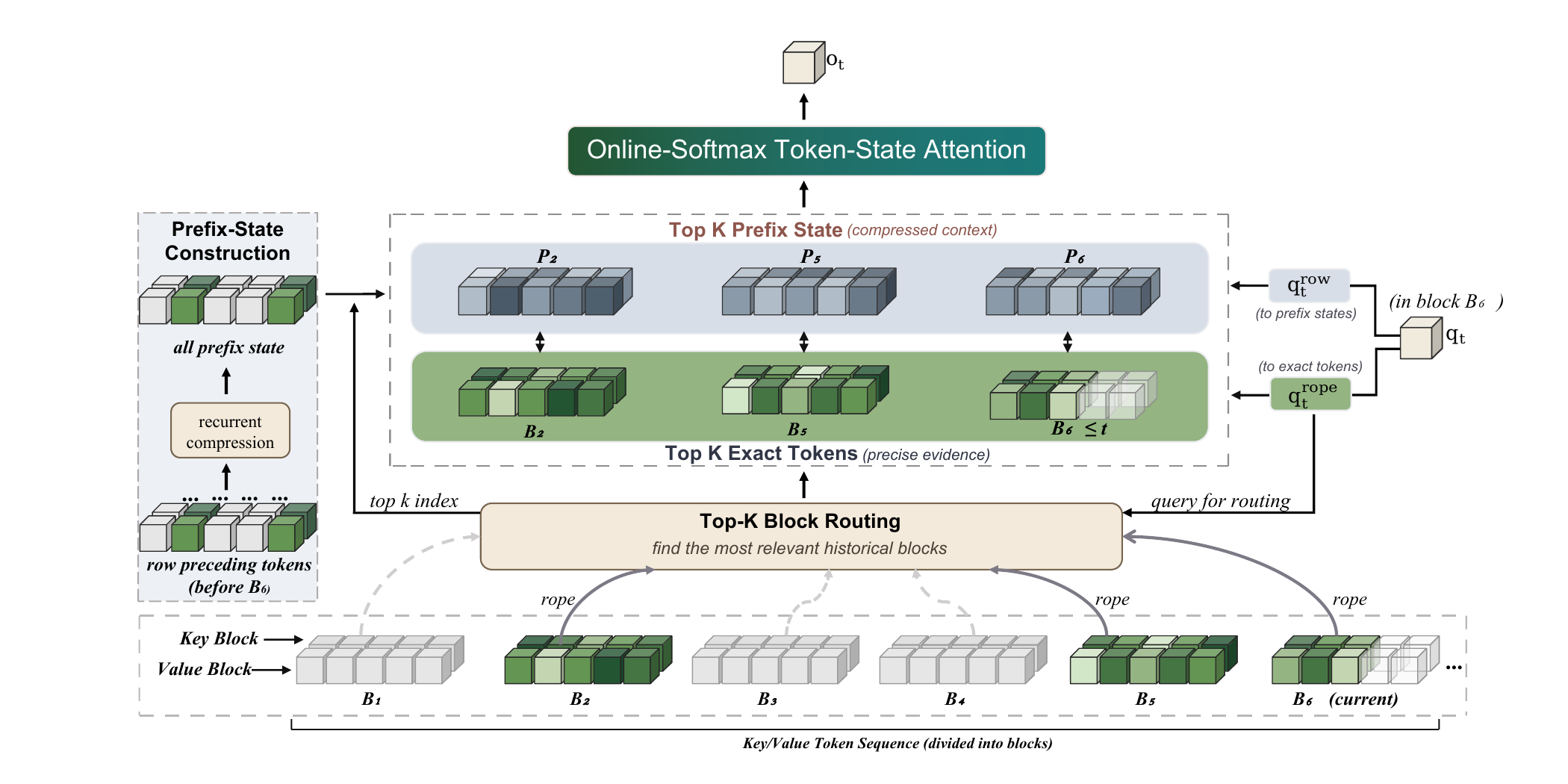}
    \caption{\small
Overview of PHBA.
The RoPE query $\bm q_t^{\mathrm{rope}}$ first performs \textbf{Top-$K$ Block Routing} to select relevant historical blocks. PHBA attends to the retrieved exact tokens with $\bm q_t^{\mathrm{rope}}$ and their \textbf{Block-Aligned Prefix States} with $\bm q_t^{\mathrm{raw}}$, and combines both memory types through \textbf{Online-Softmax Token--State Attention} to produce $\bm o_t$.
}
    \label{fig:phba_model}
\end{figure*}

Prefix-State Hybrid Block Attention (PHBA) combines block-sparse retrieval
with recurrent prefix states within a single attention layer, as illustrated
in Figure~\ref{fig:phba_model}. For each query,
PHBA retrieves a small number of relevant historical blocks at full token
resolution, while compact recurrent states summarize the prefixes preceding
these blocks. The exact token representations and their aligned prefix states
share a common softmax normalization, which is evaluated efficiently through
online-softmax merging without explicitly materializing a combined attention
matrix.
We describe PHBA in four parts: top-$K$ block routing, prefix-state construction,
online-softmax token--state attention, and efficient execution. Unless otherwise stated, we use
single-head notation. Let
$\bm{Q}^{\mathrm{raw}},\bm{K}^{\mathrm{raw}}\in\mathbb{R}^{T\times d}$
and $\bm{V}\in\mathbb{R}^{T\times d_v}$ denote the projected queries, keys,
and values. RoPE is applied to the token-attention queries and keys, producing
$\bm{Q}^{\mathrm{rope}}$ and $\bm{K}^{\mathrm{rope}}$.
\subsection{Top-$K$ Block Routing}
\label{sec:method-routing}

PHBA partitions the sequence into
$N_c=\lceil T/C\rceil$ contiguous blocks of at most $C$ tokens.
Let $\mathcal{B}_b$ denote block $b$, and let $c(t)$ denote the block
containing query token $t$. Historical routing considers only blocks preceding
the current block.

PHBA uses a total exact-block budget of $K$ blocks per query.
Since the current block is always retained for local causal attention,
$K_h=K-1$ blocks remain for historical routing.

Each block is represented for routing by the mean of its RoPE-transformed keys.
For query $t$, PHBA selects the $K_h$ historical blocks whose summaries have
the largest similarity to the query:
\begin{equation}
\bar{\bm{k}}^{\mathrm{rope}}_b
=
\frac{1}{|\mathcal{B}_b|}
\sum_{i\in\mathcal{B}_b}
\bm{k}^{\mathrm{rope}}_i,
\qquad
\mathcal{R}_t
=
\operatorname{TopK}
\left(
\left\{
(\bm q_t^{\mathrm{rope}})^\top
\bar{\bm k}^{\mathrm{rope}}_b
\right\}_{b<c(t)},
K_h
\right).
\label{eq:phba-routing}
\end{equation}

The centroid is used only for routing. Once a historical block is selected,
PHBA attends to its original token-level keys and values. The current block is not included in the historical route; instead, local
causal attention over the current block forms part of the token branch. Therefore, the exact-token memory visible to query
$t$ is
\begin{equation}
\mathcal{E}^{\mathrm{tok}}_t
=
\mathcal{B}_{c(t)}^{\le t}
\cup
\bigcup_{b\in\mathcal{R}_t}\mathcal{B}_b ,
\label{eq:phba-token-set}
\end{equation}
where
$\mathcal{B}_{c(t)}^{\le t}$ contains the causal tokens in the current block.
This provides local causal context together with query-dependent access to
distant historical blocks.

\subsection{Prefix-State Construction}
\label{sec:method-prefix-state}

Sparse retrieval provides exact token information from selected blocks,
while recurrent prefix states provide a compact summary of earlier history. PHBA
constructs separate prefix-key and prefix-value states using the gated
recurrence introduced in Sec.~\ref{sec:prelim-gated-memory}.

Let $M$ denote the number of recurrent slots. For token $t$, PHBA predicts a
retention gate
$\bm A_t=\sigma(\bm a_t)^{1/\tau}\in(0,1)^M$
and uses the coupled write gate
$\bm I_t=\bm 1-\bm A_t$.
The prefix states are updated as
\begin{equation}
\begin{aligned}
\bm S_t^K
&=
\operatorname{Diag}(\bm A_t)\bm S_{t-1}^K
+
\bm I_t(\bm k_t^{\mathrm{raw}})^\top,
\\
\bm S_t^V
&=
\operatorname{Diag}(\bm A_t)\bm S_{t-1}^V
+
\bm I_t\bm v_t^\top ,
\end{aligned}
\label{eq:phba-prefix-update}
\end{equation}
where
$\bm S_t^K\in\mathbb{R}^{M\times d}$
and
$\bm S_t^V\in\mathbb{R}^{M\times d_v}$.
The two recurrences share the same slot-wise retention and write gates, ensuring that each prefix-key slot and its corresponding prefix-value slot summarize the same historical positions with identical temporal weighting.

For a block $\mathcal{B}_b$ beginning at position $\tau_b$, PHBA stores the
states immediately before the block:
\begin{equation}
\bm P_b^K=\bm S_{\tau_b-1}^K,
\qquad
\bm P_b^V=\bm S_{\tau_b-1}^V .
\label{eq:phba-prefix-boundary}
\end{equation}
Thus, $(\bm P_b^K,\bm P_b^V)$ summarizes the causal prefix preceding
$\mathcal{B}_b$ and does not contain tokens from the block itself.

For query $t$, the candidate prefix-state blocks are
\begin{equation}
\mathcal{E}^{\mathrm{st}}_t
=
\{c(t)\}\cup\mathcal{R}_t .
\label{eq:phba-state-set}
\end{equation}
States associated with blocks having an empty causal prefix are masked.
The current-block state summarizes all history before the local block, while
each routed state provides the prefix associated with the corresponding
historical evidence block.

\subsection{Online-Softmax Token--State Attention}
\label{sec:method-phba-attn}

PHBA attends jointly to two types of memory: exact token representations and
block-aligned prefix states. The token branch uses RoPE-transformed queries and
keys, whereas the prefix-state branch uses the raw query and the raw
prefix-state keys. For a token
$i\in\mathcal{E}^{\mathrm{tok}}_t$ and a prefix-state slot $m$ from block
$b\in\mathcal{E}^{\mathrm{st}}_t$, the corresponding attention logits are
\begin{equation}
\ell^{\mathrm{tok}}_{t,i}
=
\frac{
(\bm q_t^{\mathrm{rope}})^\top
\bm k_i^{\mathrm{rope}}
}{\sqrt d},
\qquad
\ell^{\mathrm{st}}_{t,b,m}
=
\frac{
(\bm q_t^{\mathrm{raw}})^\top
\bm p^K_{b,m}
}{\sqrt d},
\label{eq:phba-logits}
\end{equation}
where
$\bm p^K_{b,m}$ and $\bm p^V_{b,m}$ denote the $m$-th key and value slots of
the prefix state associated with block $b$.

Although the two memory types are evaluated through separate sparse branches,
they are normalized jointly. Define their branch-wise partition functions as
\begin{equation}
Z_t^{\mathrm{tok}}
=
\sum_{i\in\mathcal{E}^{\mathrm{tok}}_t}
e^{\ell^{\mathrm{tok}}_{t,i}},
\qquad
Z_t^{\mathrm{st}}
=
\sum_{b\in\mathcal{E}^{\mathrm{st}}_t}
\sum_{m=1}^{M}
e^{\ell^{\mathrm{st}}_{t,b,m}} .
\label{eq:phba-partition}
\end{equation}
The PHBA output is
\begin{equation}
\bm{o}^{\mathrm{PHBA}}_t
=
\frac{
\displaystyle
\sum_{i\in\mathcal{E}^{\mathrm{tok}}_t}
e^{\ell^{\mathrm{tok}}_{t,i}}\bm v_i
+
\displaystyle
\sum_{b\in\mathcal{E}^{\mathrm{st}}_t}
\sum_{m=1}^{M}
e^{\ell^{\mathrm{st}}_{t,b,m}}\bm p^V_{b,m}
}{
Z_t^{\mathrm{tok}}+Z_t^{\mathrm{st}}
}.
\label{eq:phba-output}
\end{equation}

Importantly, PHBA does not explicitly concatenate the token and prefix-state
memories. Instead, each branch maintains its own online-softmax statistics.
Let
$\bm O_t^{\mathrm{tok}},L_t^{\mathrm{tok}}$
and
$\bm O_t^{\mathrm{st}},L_t^{\mathrm{st}}$
denote the normalized outputs and log-sum-exp statistics of the token and
prefix-state branches, respectively. The two branches are merged exactly as
\begin{equation}
\begin{aligned}
L_t
&=
\operatorname{LSE}
\left(
L_t^{\mathrm{tok}},
L_t^{\mathrm{st}}
\right),\\
\bm O_t
&=
e^{L_t^{\mathrm{tok}}-L_t}
\bm O_t^{\mathrm{tok}}
+
e^{L_t^{\mathrm{st}}-L_t}
\bm O_t^{\mathrm{st}} .
\end{aligned}
\label{eq:phba-online-merge}
\end{equation}

This online-softmax merge is exactly equivalent to applying a single softmax
over the union of exact tokens and prefix-state slots, while avoiding explicit
concatenation of the two memory types. Consequently, each query allocates
probability mass directly between exact token evidence and compressed prefix
information, rather than combining independently normalized branch outputs
through an external fusion weight.
In implementation, the two branch outputs need not be materialized separately:
the same online-softmax merge is applied incrementally while streaming the
selected token and prefix-state blocks.
The hardware-efficient realization is described in
Sec.~\ref{sec:method-chunkwise} and
Algorithm~\ref{alg:phba-app-sparse-forward}.

\subsection{Hardware-Efficient Chunkwise and Sparse Execution}
\label{sec:method-chunkwise}

PHBA addresses two sources of training inefficiency.
First, the recurrent prefix-state construction is reformulated into a
chunkwise-parallel computation, avoiding token-wise recurrent execution within
each chunk. Second, routed attention is implemented with metadata-driven sparse
kernels that avoid gathering selected queries and memory blocks into packed
payload tensors. Together, these two designs enable efficient training while
preserving the PHBA formulation in
Sec.~\ref{sec:method-phba-attn}.

\paragraph{Chunkwise Prefix-State Update.}
The prefix-state recurrence follows the gated slot-memory structure of GSA
and admits the same chunkwise decomposition as gated linear attention.
We use the same block partition as in Sec.~\ref{sec:method-routing}.
For packed inputs, the recurrence is applied independently to each sequence
and the incoming state is reset to zero at its first block.
Let $C_n=|\mathcal{B}_n|$ denote the length of block $n$.
For block $[n]$, define
\begin{equation}
\overrightarrow{\bm A}_{[n],i}
=
\prod_{j=1}^{i}\bm A_{[n],j},
\qquad
\overleftarrow{\bm A}_{[n],i}
=
\prod_{j=i+1}^{C_n}\bm A_{[n],j}.
\end{equation}
where the products are elementwise. For compact notation, we write
$\bm S_{[n]}=[\bm S^K_{[n]},\bm S^V_{[n]}]$ and
$\bm Z_{[n]}=[\bm K^{\mathrm{raw}}_{[n]},\bm V_{[n]}]$.
Let
$\bm I_{[n]}=\bm 1-\bm A_{[n]}$ and
$\bm W_{[n]}
=
\bm I_{[n]}\odot\overleftarrow{\bm A}_{[n]}$.
The boundary state is updated as
\begin{equation}
\bm S_{[n+1]}
=
\operatorname{Diag}
\left(
\overrightarrow{\bm A}_{[n],C_n}
\right)
\bm S_{[n]}
+
\bm W_{[n]}^\top\bm Z_{[n]} .
\label{eq:phba-chunk-state}
\end{equation}

Only the boundary state is recurrent across chunks, while the writes within
each chunk are evaluated in parallel. The incoming state $\bm S_{[n]}$ is
stored as the prefix state associated with chunk $[n]$. Thus, PHBA retains the
causal recurrent update while exposing only block-aligned prefix states to the
subsequent sparse attention computation.
A detailed derivation of Eq.~\ref{eq:phba-chunk-state} is provided in
Appendix~\ref{sec:appendix-phba-state-kernel}.

\paragraph{Operator-Aware Sparse Execution.}
PHBA is designed around a simple systems constraint: routed attention should not
first gather selected query rows and memory blocks into packed sparse payload
tensors. Such a pipeline would introduce additional HBM round trips before the
attention kernel can start. PHBA instead materializes only compact integer
routing metadata and directly loads the required query rows, token blocks, and
prefix-state blocks from their original HBM layouts into on-chip
SRAM/registers.

Algorithm~\ref{alg:phba-routing-work} first produces the query-centric
historical routes used by the forward pass. During training, the same routing
decisions are additionally reformatted into block-major variable-length
metadata for the backward pass, without moving the underlying payloads.
For the forward pass, block-local causal FlashAttention first computes
row-wise outputs and log-sum-exp statistics. The fused sparse kernel then
extends the corresponding online-softmax accumulators with the current-block
prefix state, routed historical token blocks, and their aligned prefix states.
This is exactly equivalent to the joint token--state softmax in
Eq.~\ref{eq:phba-output} and the online-softmax merge in
Eq.~\ref{eq:phba-online-merge}, without explicitly concatenating the two
memory types. Algorithm~\ref{alg:phba-forward} summarizes the forward
execution. Detailed centroid computation, top-$K$ selection, metadata
reformatting, sparse forward, and backward kernels are provided in
Appendix~\ref{sec:appendix-phba-kernels}.
\begin{algorithm}[t]
\caption{PHBA Routing and Block-Major Metadata Construction}
\label{alg:phba-routing-work}
\begin{algorithmic}[1]

\Require RoPE queries/keys
$\bm{Q}^{\mathrm{rope}},\bm{K}^{\mathrm{rope}}
\in\mathbb{R}^{T\times H\times d}$ in HBM.
Boundaries $\mathrm{cu}$. Block size $C$.
Total exact-block budget $K$.

\State Set historical routing budget $K_h\leftarrow K-1$.

\State Construct current block ids
$\bm{c}\in\mathbb{Z}^{T}$,
block-row pointers
$\mathrm{BlockPtr}\in\mathbb{Z}^{N_c+1}$,
and prefix-validity mask
$\mathrm{PrefixValid}\in\{0,1\}^{N_c}$
from $\mathrm{cu}$ and $C$, where
$\mathrm{PrefixValid}[b]=1$ iff block $b$
has a nonempty causal prefix.

\State Compute key-block centroids
$\bar{\bm{K}}^{\mathrm{rope}}
\in\mathbb{R}^{N_c\times H\times d}$
using Algorithm~\ref{alg:phba-app-centroid}.

\State Select historical top-$K_h$ blocks
$\bm{R}^{c}\in\mathbb{Z}^{T\times H\times K_h}$
using Algorithm~\ref{alg:phba-app-topk}.

\State Set historical-token routes
$\bm{R}^{\rm tok}\leftarrow\bm{R}^{c}$.

\State Broadcast $\bm{c}$ over heads as
$\bm{c}^{H}\in\mathbb{Z}^{T\times H\times 1}$
and set prefix-state routes
$\bm{R}^{\rm st}\leftarrow[\bm{c}^{H};\bm{R}^{c}]$
along the route dimension.

\State Set any entry in $\bm{R}^{\rm st}$ whose routed block $b$
has $\mathrm{PrefixValid}[b]=0$ to $-1$.

\State During training, reformat
$\bm{R}^{\rm tok}$ and $\bm{R}^{\rm st}$ into
block-major varlen metadata for sparse backward using
Algorithm~\ref{alg:phba-app-sparse-metadata}.

\State \Return
$\bm{R}^{c},\bm{c},\mathrm{BlockPtr},\mathrm{PrefixValid}$ and,
during training,
$(\bm{\rho}^{\rm tok},\mathrm{Offset}^{\rm tok},
C_{\rm blk}^{\rm tok})$,
$(\bm{\rho}^{\rm st},\mathrm{Offset}^{\rm st},
C_{\rm blk}^{\rm st})$.

\end{algorithmic}
\end{algorithm}

\begin{algorithm}[t]
\caption{PHBA Forward Pass}
\label{alg:phba-forward}
\begin{algorithmic}[1]

\Require Token tensors
$\bm{Q}^{\mathrm{raw}},\bm{Q}^{\mathrm{rope}},
\bm{K}^{\mathrm{rope}}
\in\mathbb{R}^{T\times H\times d}$
and
$\bm{V}\in\mathbb{R}^{T\times H\times d_v}$ in HBM.
Prefix states
$\bm{P}^{K}\in\mathbb{R}^{N_c\times H\times M\times d}$
and
$\bm{P}^{V}\in\mathbb{R}^{N_c\times H\times M\times d_v}$
in HBM.
Routes $\bm{R}^{c}$, current block ids $\bm{c}$,
block-row pointers $\mathrm{BlockPtr}$,
and prefix-validity mask $\mathrm{PrefixValid}$.
Tile sizes $B_r,B_c$.

\State Run block-local causal FlashAttention according to
$\mathrm{BlockPtr}$ and write
$\bm{O}^{\rm loc}\in\mathbb{R}^{T\times H\times d_v}$
and
$L^{\rm loc}\in\mathbb{R}^{T\times H}$
to HBM.

\State Run \textsc{FusedSparseForward}
(Algorithm~\ref{alg:phba-app-sparse-forward}) with
$(\bm{Q}^{\mathrm{raw}},\bm{Q}^{\mathrm{rope}},
\bm{K}^{\mathrm{rope}},\bm{V},
\bm{P}^{K},\bm{P}^{V})$,
routes $(\bm{R}^{c},\bm{c})$,
$\mathrm{BlockPtr}$,
$\mathrm{PrefixValid}$, seed
$(\bm{O}^{\rm loc},L^{\rm loc})$,
and tile sizes $B_r,B_c$.

\State The fused kernel extends the corresponding online-softmax
accumulators with the current-block prefix state,
historical exact-token blocks, and their aligned prefix states.

\State The fused kernel writes final output $\bm{O}$
and joint log-sum-exp $L$ to HBM.

\State \Return $\bm{O},L$.

\end{algorithmic}
\end{algorithm}

\section{Experiments}

\subsection{Experimental Setup}
\label{sec:exp-setup}

\paragraph{Models and Training.}
We evaluate PHBA at two scales: 760M models trained on 50B tokens and
1.3B models trained on 100B tokens. All models are trained from scratch on
FineWeb-Edu~\citep{penedo2024fineweb} with an 8K context length. Unless otherwise specified, PHBA uses
$M=128$ prefix-state slots and block size $C=128$.
For the 1.3B main comparison, we use a fixed $1/8$ exact-token budget,
corresponding to $K=8$ blocks at 8K.
Additional architecture, optimization, and training details are provided in
Appendix~\ref{sec:m-a-t}.

\paragraph{Baselines.}
We compare PHBA with Transformer~\citep{vaswani2017attention},
MoBA~\citep{lu2026moba}, GSA~\citep{zhang2024gated},
KDA~\citep{team2025kimi}, and NHA~\citep{du2026native}.
All baselines are trained on the same data and context length at matched model
scale; efficient baselines use matched token budgets where applicable. Detailed baseline information is provided in Appendix~\ref{sec:b-c}.

\paragraph{Evaluation.}
We evaluate language modeling, zero-shot commonsense reasoning, long-context
retrieval, and real-world understanding. Language modeling and commonsense
evaluation use \texttt{lm-evaluation-harness}~\citep{eval-harness}.
Models trained at 8K are directly evaluated at 16K--64K without further
training. We use RULER/NIAH for long-context retrieval and LongBench for
real-world understanding, truncating LongBench inputs beyond 8K to the first
and last 4K tokens. Detailed benchmark descriptions are provided in
Appendix~\ref{sec:e-p}.
\subsection{Short-Context Language Modeling and Retrieval}
\label{sec:exp-short-context}

Table~\ref{tab:comprehensive_merged_results_fixed} reports performance within
the 8K training context.

\paragraph{Language Modeling and Commonsense Reasoning.}
Table~\ref{tab:comprehensive_merged_results_fixed} reports WikiText and
LAMBADA~\citep{merity2016pointer,paperno2016lambada} together with zero-shot
ARC, HellaSwag, PIQA, and WinoGrande
~\citep{clark2018think,zellers2019hellaswag,bisk2020piqa,sakaguchi2021winogrande}.
At 1.3B scale, PHBA achieves the best WikiText and LAMBADA results among the
compared models and a competitive commonsense average of 60.07, second only
to KDA. This shows that routed retrieval and prefix-state memory preserve
strong short-context modeling quality.
\begin{table*}[!t]
    \centering
    \caption{
\small
\textbf{Short-context evaluation of 760M and 1.3B models.}
RULER/NIAH are evaluated at 4K and 8K and scaled by 100.
\textbf{Bold}/\underline{underline} denote best/second-best results.
PHBA (default): $M=C=128$, $K=8$.
}
    \vspace{-0.25em}
    \begin{adjustbox}{width=1.0\textwidth, center}
    \normalsize
    \renewcommand{\arraystretch}{0.95} 
    \renewcommand{\multirowsetup}{\centering}
    \setlength{\tabcolsep}{1.5pt}
    \begin{threeparttable}

    \begin{tabular}{ 
        p{3.5cm} | 
        *{6}{>{\centering\arraybackslash}p{1.00cm}} | 
        *{12}{>{\centering\arraybackslash}p{0.78cm}} 
    }
        \toprule
        \multicolumn{19}{c}{\makebox[0pt]{\textbf{PART I: Common Sense Reasoning \& RULER Tasks}}} \\
        \midrule

        \multirow{3}{*}{Model \& Scale} 
        & \multicolumn{6}{c|}{\textbf{Common Sense Reasoning}} 
        & \multicolumn{12}{c}{\textbf{RULER Tasks}} \\

        \cmidrule(lr){2-7} 
        \cmidrule(lr){8-19}

        & ARC$_e$ 
        & ARC$_c$ 
        & Hella. 
        & PIQA 
        & Wino. 
        & Avg. 
        & \multicolumn{2}{c}{CWE} 
        & \multicolumn{2}{c}{FWE} 
        & \multicolumn{2}{c}{HotpotQA} 
        & \multicolumn{2}{c}{SQuAD} 
        & \multicolumn{2}{c|}{VT} 
        & \multicolumn{2}{c}{\textbf{Avg}} \\

        & (acc$_\uparrow$) 
        & (acc$_\uparrow$) 
        & (acc$_\uparrow$) 
        & (acc$_\uparrow$) 
        & (acc$_\uparrow$) 
        & (acc$_\uparrow$) 
        & 4k$_\uparrow$ 
        & 8k$_\uparrow$ 
        & 4k$_\uparrow$ 
        & 8k$_\uparrow$ 
        & 4k$_\uparrow$ 
        & 8k$_\uparrow$ 
        & 4k$_\uparrow$ 
        & 8k$_\uparrow$ 
        & 4k$_\uparrow$ 
        & 8k$_\uparrow$ 
        & 4k$_\uparrow$ 
        & 8k$_\uparrow$ \\

        \midrule

        \multicolumn{19}{l}{\textit{760M params with 50B training tokens}} \\

        \rowcolor{blue!10} 
        PHBA (origin) 
        & \underline{66.96} 
        & \underline{34.30} 
        & \underline{52.14} 
        & \underline{70.73} 
        & 55.33 
        & \cellcolor{red!7}\underline{55.89} 
        & 26.3 & 3.6 
        & \textbf{19.5} & \textbf{22.1} 
        & 20.2 & 15.6 
        & 24.4 & 14.6 
        & 1.0 & \textbf{2.0} 
        & \cellcolor{red!7}\underline{18.3} 
        & \cellcolor{red!7}\underline{11.6} \\

        ~~w. $C=512, K=2$ 
        & 64.10 & 32.34 & 48.25 & 69.70 & 52.25 
        & \cellcolor{red!7}53.33 
        & 4.8 & 0.3 
        & 1.6 & 3.4 
        & 23.0 & 16.4 
        & 26.6 & 10.0 
        & \textbf{10.9} & 0.6 
        & \cellcolor{red!7}13.4 
        & \cellcolor{red!7}6.1 \\

        ~~w. $C=256, K=4$ 
        & 64.35 & 32.00 & 49.89 & 69.70 & 53.67 
        & \cellcolor{red!7}53.92 
        & 20.0 & 0.6 
        & 4.1 & 9.2 
        & 21.0 & 15.8 
        & 24.3 & 9.7 
        & \underline{9.0} & 0.4 
        & \cellcolor{red!7}15.7 
        & \cellcolor{red!7}7.1 \\

        ~~w. $C=64, K=16$ 
        & \textbf{67.51} & 34.22 
        & \textbf{53.25} 
        & \textbf{71.38} 
        & 55.88 
        & \cellcolor{red!7}\textbf{56.45} 
        & \textbf{34.7} & \textbf{5.5} 
        & 11.0 & 3.9 
        & \underline{24.4} & \textbf{22.2} 
        & \textbf{37.8} & \textbf{22.1} 
        & 0.9 & 0.3 
        & \cellcolor{red!7}\textbf{21.8} 
        & \cellcolor{red!7}10.8 \\

        ~~w. $M=64$ 
        & 66.08 & 32.59 & 51.70 & 70.40 & 55.64 
        & \cellcolor{red!7}55.28 
        & 10.1 & 3.1 
        & 2.9 & 2.7 
        & \textbf{24.6} & \underline{19.0} 
        & \underline{33.1} & \underline{19.6} 
        & 1.8 & 0.0 
        & \cellcolor{red!7}14.5 
        & \cellcolor{red!7}8.9 \\

        ~~w. $M=256$ 
        & 64.94 & 31.31 & 48.85 & 70.13 & \underline{55.96} 
        & \cellcolor{red!7}54.24 
        & 21.4 & \underline{4.0} 
        & 0.0 & 0.0 
        & 23.4 & 18.8 
        & 27.9 & 12.9 
        & 4.8 & 0.1 
        & \cellcolor{red!7}15.5 
        & \cellcolor{red!7}7.2 \\

        ~~w/o. gate 
        & 64.52 & \textbf{34.39} & 49.98 & 70.24 & 53.75 
        & \cellcolor{red!7}54.58 
        & \underline{30.5} & 1.1 
        & 4.1 & 2.1 
        & 20.0 & 17.0 
        & 21.9 & 10.9 
        & 0.6 & 0.0 
        & \cellcolor{red!7}15.4 
        & \cellcolor{red!7}6.2 \\

        ~~w. mean 
        & 65.61 & 33.53 & 50.19 & 70.24 & \textbf{57.38} 
        & \cellcolor{red!7}55.39 
        & 18.1 & 0.1 
        & 7.5 & 4.1 
        & 20.8 & 17.8 
        & 24.9 & 15.2 
        & 7.4 & \underline{0.8} 
        & \cellcolor{red!7}15.7 
        & \cellcolor{red!7}7.6 \\

        \midrule

        \multicolumn{19}{l}{\textit{1.3B params with 100B training tokens}} \\

        \rowcolor{gray!30} 
        Transformer 
        & \underline{72.14} 
        & \underline{38.74} 
        & 57.41 
        & 71.33 
        & 58.33 
        & \cellcolor{red!7}59.59 
        & \textbf{38.8} & 2.2 
        & 25.7 & 27.2 
        & \textbf{30.2} & \underline{24.6}
        & 34.3 & \underline{25.6} 
        & 5.6 & 0.0 
        & \cellcolor{red!7}\underline{26.9} 
        & \cellcolor{red!7}15.9 \\

        MoBA 
        & 70.16 & 36.09 & 56.25 & 71.38 & 55.96 
        & \cellcolor{red!7}57.97 
        & \underline{35.2} & 1.6 
        & 18.2 & 3.9 
        & 25.2 & 21.0 
        & \underline{37.4} & 22.0 
        & 0.8 & 5.5 
        & \cellcolor{red!7}23.4 
        & \cellcolor{red!7}10.8 \\

        GSA 
        & 68.94 & 35.84 & 56.15 & 72.31 & 56.67 
        & \cellcolor{red!7}57.98 
        & 1.5 & 0.4 
        & 1.3 & 2.0 
        & 17.2 & 15.8 
        & 14.4 & 10.1 
        & \textbf{32.4} & \textbf{23.6} 
        & \cellcolor{red!7}13.4 
        & \cellcolor{red!7}10.4 \\

        KDA 
        & \textbf{73.53} 
        & \textbf{40.02} 
        & \textbf{58.89} 
        & \underline{73.45} 
        & \underline{59.43} 
        & \cellcolor{red!7}\textbf{61.06} 
        & 28.6 & 3.5 
        & \underline{41.8} & \underline{33.3} 
        & 23.6 & 21.6 
        & 23.4 & 14.1 
        & 7.6 & 10.8 
        & \cellcolor{red!7}25.0 
        & \cellcolor{red!7}\underline{16.7} \\

        NHA
        & 71.42 
        & 36.09 
        & 58.35 
        & \textbf{74.37} 
        & 58.48 
        & \cellcolor{red!7}59.75
        & 33.5 & \underline{7.0}
        & \textbf{45.7} & 23.1
        & 11.8 & \textbf{26.2}
        & 19.2 & \textbf{27.2}
        & \underline{21.8} & 1.3
        & \cellcolor{red!7}26.4
        & \cellcolor{red!7}15.1 \\

        \rowcolor{blue!10} 
        PHBA 
        & 71.17 
        & 38.14 
        & \underline{58.58} 
        & 72.63 
        & \textbf{59.83} 
        & \cellcolor{red!7}\underline{60.07} 
        & 31.0 & \textbf{24.2}
        & 31.3 & \textbf{45.3}
        & \underline{29.0} & 13.4
        & \textbf{43.4} & 16.2
        & 1.5 & \underline{20.5}
        & \cellcolor{red!7}\textbf{27.2}
        & \cellcolor{red!7}\textbf{23.9} \\

        % ============================================================
        % PART II
        % ============================================================
        \midrule\midrule

        \multicolumn{19}{c}{
            \makebox[0pt]{\textbf{PART II: Language Modeling Tasks}}
        } \\

        \midrule

        \multicolumn{7}{c|}{
            \textbf{760M params with 50B training tokens}
        } 
        & \multicolumn{12}{c}{
            \textbf{1.3B params with 100B training tokens}
        } \\

        \cmidrule(lr){1-7} 
        \cmidrule(lr){8-19}

        Model 
        & \multicolumn{2}{c}{Wiki. (ppl$_\downarrow$)}
        & \multicolumn{2}{c}{Lamb. (acc$_\uparrow$)}
        & \multicolumn{2}{c|}{Lamb. (ppl$_\downarrow$)}

        & \multicolumn{3}{l|}{Model}
        & \multicolumn{3}{c}{Wiki. (ppl$_\downarrow$)}
        & \multicolumn{3}{c}{Lamb. (acc$_\uparrow$)}
        & \multicolumn{3}{c}{Lamb. (ppl$_\downarrow$)} \\

        \midrule

        % Row 1
        \cellcolor{blue!10}PHBA (origin)
        & \multicolumn{2}{c}{\cellcolor{blue!10}\underline{18.33}}
        & \multicolumn{2}{c}{\cellcolor{blue!10}42.98}
        & \multicolumn{2}{c|}{\cellcolor{blue!10}16.33}

        & \multicolumn{3}{l|}{\cellcolor{gray!30}Transformer}
        & \multicolumn{3}{c}{\cellcolor{gray!30}\underline{15.67}}
        & \multicolumn{3}{c}{\cellcolor{gray!30}46.69}
        & \multicolumn{3}{c}{\cellcolor{gray!30}12.26} \\

        % Row 2
        ~~w. $C=512, K=2$
        & \multicolumn{2}{c}{19.76}
        & \multicolumn{2}{c}{40.31}
        & \multicolumn{2}{c|}{19.34}

        & \multicolumn{3}{l|}{MoBA}
        & \multicolumn{3}{c}{16.22}
        & \multicolumn{3}{c}{47.22}
        & \multicolumn{3}{c}{13.04} \\

        % Row 3
        ~~w. $C=256, K=4$
        & \multicolumn{2}{c}{19.19}
        & \multicolumn{2}{c}{40.52}
        & \multicolumn{2}{c|}{17.86}

        & \multicolumn{3}{l|}{GSA}
        & \multicolumn{3}{c}{16.86}
        & \multicolumn{3}{c}{40.93}
        & \multicolumn{3}{c}{16.74} \\

        % Row 4
        ~~w. $C=64, K=16$
        & \multicolumn{2}{c}{\textbf{17.34}}
        & \multicolumn{2}{c}{\textbf{45.24}}
        & \multicolumn{2}{c|}{\textbf{14.60}}

        & \multicolumn{3}{l|}{KDA}
        & \multicolumn{3}{c}{16.18}
        & \multicolumn{3}{c}{\underline{48.65}}
        & \multicolumn{3}{c}{\underline{10.91}} \\

        % Row 5
        ~~w. $M=64$
        & \multicolumn{2}{c}{18.84}
        & \multicolumn{2}{c}{\underline{43.74}}
        & \multicolumn{2}{c|}{\underline{15.91}}

        & \multicolumn{3}{l|}{NHA}
        & \multicolumn{3}{c}{15.96}
        & \multicolumn{3}{c}{48.34}
        & \multicolumn{3}{c}{11.51} \\

        % Row 6
        ~~w. $M=256$
        & \multicolumn{2}{c}{19.74}
        & \multicolumn{2}{c}{41.90}
        & \multicolumn{2}{c|}{18.15}

        & \multicolumn{3}{l|}{\cellcolor{blue!10}PHBA}
        & \multicolumn{3}{c}{\cellcolor{blue!10}\textbf{15.21}}
        & \multicolumn{3}{c}{\cellcolor{blue!10}\textbf{49.27}}
        & \multicolumn{3}{c}{\cellcolor{blue!10}\textbf{10.80}} \\

        % Row 7
        ~~w/o. gate
        & \multicolumn{2}{c}{18.88}
        & \multicolumn{2}{c}{43.31}
        & \multicolumn{2}{c|}{16.11}

        & \multicolumn{12}{c}{\textbf{Table Color Legend}} \\

        % Row 8
        ~~w. mean
        & \multicolumn{2}{c}{18.76}
        & \multicolumn{2}{c}{43.24}
        & \multicolumn{2}{c|}{16.38}

        & \multicolumn{4}{c}{
            \cellcolor{blue!10}\textbf{Blue:} Proposed Method
        }
        & \multicolumn{4}{c}{
            \cellcolor{gray!30}\textbf{Gray:} Primary Baseline
        }
        & \multicolumn{4}{c}{
            \cellcolor{red!7}\textbf{Red:} Average Columns
        } \\

        % ============================================================
        % PART III
        % ============================================================
        \midrule\midrule

        \multicolumn{19}{c}{
            \makebox[0pt]{\textbf{PART III: NIAH Tasks}}
        } \\

        \midrule

        \multirow{3}{*}{Model \& Scale}
        & \multicolumn{6}{c|}{\textbf{Single-Key NIAH}}
        & \multicolumn{6}{c|}{\textbf{Multi-Key NIAH}}
        & \multicolumn{4}{c|}{\textbf{NIAH (Extra)}}
        & \multicolumn{2}{c}{\textbf{Avg}} \\

        \cmidrule(lr){2-7} 
        \cmidrule(lr){8-13} 
        \cmidrule(lr){14-17} 
        \cmidrule(lr){18-19}

        & \multicolumn{2}{c}{Task 1}
        & \multicolumn{2}{c}{Task 2}
        & \multicolumn{2}{c|}{Task 3}
        & \multicolumn{2}{c}{Task 1}
        & \multicolumn{2}{c}{Task 2}
        & \multicolumn{2}{c|}{Task 3}
        & \multicolumn{2}{c}{MultiQuery}
        & \multicolumn{2}{c|}{MultiValue}
        & \multicolumn{2}{c}{Mean} \\

        & 4k$_\uparrow$ & 8k$_\uparrow$
        & 4k$_\uparrow$ & 8k$_\uparrow$
        & 4k$_\uparrow$ & 8k$_\uparrow$
        & 4k$_\uparrow$ & 8k$_\uparrow$
        & 4k$_\uparrow$ & 8k$_\uparrow$
        & 4k$_\uparrow$ & 8k$_\uparrow$
        & 4k$_\uparrow$ & 8k$_\uparrow$
        & 4k$_\uparrow$ & 8k$_\uparrow$
        & 4k$_\uparrow$ & 8k$_\uparrow$ \\

        \midrule

        \multicolumn{19}{l}{
            \textit{760M params with 50B training tokens}
        } \\

        \rowcolor{blue!10} 
        PHBA (origin)
        & \textbf{99.4} & 20.4
        & \underline{97.4} & 42.8
        & \textbf{67.4} & 11.4
        & \underline{78.0} & \underline{44.0}
        & \textbf{43.2} & \underline{5.6}
        & \textbf{18.8} & 0.0
        & \underline{41.7} & \underline{22.2}
        & \textbf{46.0} & \underline{25.0}
        & \cellcolor{red!7}\textbf{61.5}
        & \cellcolor{red!7}21.4 \\

        ~~w. $C=512, K=2$
        & 78.6 & 12.6
        & 36.2 & 19.6
        & 25.4 & 3.6
        & 32.0 & 20.4
        & 1.4 & 0.6
        & 1.0 & 0.2
        & 23.1 & 13.9
        & 23.1 & 15.8
        & \cellcolor{red!7}27.6
        & \cellcolor{red!7}10.8 \\

        ~~w. $C=256, K=4$
        & 88.0 & 6.2
        & 66.6 & 33.2
        & 35.2 & 6.0
        & 43.2 & 26.0
        & 5.6 & 2.4
        & \underline{12.6} & \underline{0.6}
        & 27.1 & 19.0
        & 29.3 & 20.3
        & \cellcolor{red!7}38.5
        & \cellcolor{red!7}14.2 \\

        ~~w. $C=64, K=16$
        & \underline{99.2} & \underline{52.0}
        & \textbf{98.8} & \textbf{61.6}
        & \underline{57.6} & \textbf{12.2}
        & \textbf{82.8} & \textbf{51.8}
        & \underline{24.2} & \textbf{5.8}
        & 11.6 & \textbf{1.6}
        & \textbf{44.8} & 20.8
        & \underline{42.9} & \textbf{25.3}
        & \cellcolor{red!7}\underline{57.7}
        & \cellcolor{red!7}\textbf{28.9} \\

        ~~w. $M=64$
        & 99.0 & 20.2
        & 93.4 & \underline{50.4}
        & 41.4 & 12.0
        & 72.2 & 42.2
        & 10.2 & 1.4
        & 9.4 & 0.2
        & 32.7 & \textbf{23.0}
        & 39.0 & 23.2
        & \cellcolor{red!7}49.7
        & \cellcolor{red!7}\underline{21.6} \\

        ~~w. $M=256$
        & 66.2 & 3.6
        & 75.4 & 23.2
        & 13.2 & 4.6
        & 42.6 & 18.4
        & 2.6 & 0.2
        & 7.6 & 0.0
        & 29.7 & 9.3
        & 31.4 & 11.9
        & \cellcolor{red!7}33.6
        & \cellcolor{red!7}8.9 \\

        ~~w/o. gate
        & 92.6 & 37.4
        & 65.2 & 21.0
        & 32.8 & 5.6
        & 36.0 & 16.2
        & 2.6 & 0.6
        & 4.8 & 0.4
        & 30.5 & 13.5
        & 24.9 & 14.0
        & \cellcolor{red!7}36.2
        & \cellcolor{red!7}13.6 \\

        ~~w. mean
        & 94.8 & \textbf{53.8}
        & 68.6 & 36.6
        & 4.2 & 6.0
        & 50.0 & 35.6
        & 9.0 & 0.4
        & 4.4 & 0.0
        & 29.4 & 17.3
        & 29.9 & 20.8
        & \cellcolor{red!7}36.3
        & \cellcolor{red!7}21.3 \\

        \midrule

        \multicolumn{19}{l}{
            \textit{1.3B params with 100B training tokens}
        } \\

        \rowcolor{gray!30} 
        Transformer
        & \underline{99.8} & 72.4
        & \textbf{100.0} & \textbf{73.6}
        & \textbf{63.2} & \textbf{42.6}
        & \textbf{84.2} & \textbf{68.0}
        & \textbf{75.8} & \textbf{40.2}
        & \underline{12.0} & \textbf{1.0}
        & \textbf{57.9} & \textbf{31.3}
        & 43.2 & \underline{26.8}
        & \cellcolor{red!7}\textbf{67.0}
        & \cellcolor{red!7}\textbf{44.5} \\

        MoBA
        & 98.8 & 64.0
        & 83.0 & 37.4
        & 15.8 & 10.6
        & 41.8 & 32.8
        & 7.2 & 3.2
        & 8.6 & 0.0
        & 30.0 & 19.0
        & 29.5 & 21.4
        & \cellcolor{red!7}39.3
        & \cellcolor{red!7}23.6 \\

        GSA
        & 93.0 & 60.8
        & 88.0 & 20.2
        & 2.4 & 1.0
        & 24.2 & 21.6
        & 0.0 & 0.0
        & 0.0 & 0.0
        & 34.4 & 14.2
        & 34.4 & 14.0
        & \cellcolor{red!7}34.6
        & \cellcolor{red!7}16.5 \\

        KDA
        & \textbf{100.0} & 29.6
        & 86.2 & \underline{49.2}
        & 46.0 & 19.4
        & 26.4 & \underline{53.0}
        & 0.2 & \underline{14.0}
        & 0.0 & \underline{0.4}
        & 42.7 & \underline{27.1}
        & \underline{49.3} & \textbf{28.0}
        & \cellcolor{red!7}43.8
        & \cellcolor{red!7}27.5 \\

        NHA
        & \textbf{100.0} & \underline{98.6}
        & 95.0 & 48.2
        & 53.8 & 10.8
        & 30.0 & 24.8
        & 40.0 & 0.0
        & 0.0 & 0.0
        & 38.2 & 19.4
        & 39.1 & 17.4
        & \cellcolor{red!7}49.5
        & \cellcolor{red!7}27.4 \\

        \rowcolor{blue!10} 
        PHBA
        & \textbf{100.0} & \textbf{99.4}
        & \underline{98.6} & 39.6
        & \underline{54.4} & \underline{23.6}
        & \underline{76.0} & 26.6
        & \underline{45.8} & 0.0
        & \textbf{13.0} & 0.0
        & \underline{55.8} & 20.5
        & \textbf{59.0} & 26.1
        & \cellcolor{red!7}\underline{62.6}
        & \cellcolor{red!7}\underline{29.5} \\

        \bottomrule
    \end{tabular}

    \end{threeparttable}
    \end{adjustbox}

    \label{tab:comprehensive_merged_results_fixed}
\end{table*}
\paragraph{Short-Context Retrieval.}
On RULER~\citep{hsieh2024ruler} and its NIAH variants, PHBA consistently
outperforms the other efficient baselines at 4K and 8K, trailing only the
dense Transformer on NIAH.
This demonstrates strong retrieval within the training context.

\subsection{Long-Context Retrieval and Understanding}
\label{sec:exp-long-context}

We next evaluate the 1.3B models beyond the 8K training length.
Figure~\ref{fig:phba_results} (a)--(c) reports RULER, NIAH, and LongBench
averages, with complete results in
Appendix Table~\ref{tab:reordered_long_context_final}.

\paragraph{Long-Context Retrieval.}
PHBA maintains the strongest overall RULER and NIAH~\citep{hsieh2024ruler} performance from 16K to
64K, with the advantage becoming more pronounced beyond the training length.
Content-dependent routing preserves direct access to distant token-level
information, while aligned prefix states provide the context preceding the
retrieved blocks, improving long-context extrapolation over compressed memory
or sparse token retrieval alone.

\paragraph{Long-Context Understanding.}
PHBA also achieves the strongest LongBench~\citep{bai2024longbench} average, with gains across document
QA, summarization, and few-shot tasks while remaining competitive on code.
This indicates that the retrieval gains transfer beyond controlled synthetic
settings to real-world long-context tasks.
\begin{figure*}[t]
    \centering
    \includegraphics[width=\textwidth]{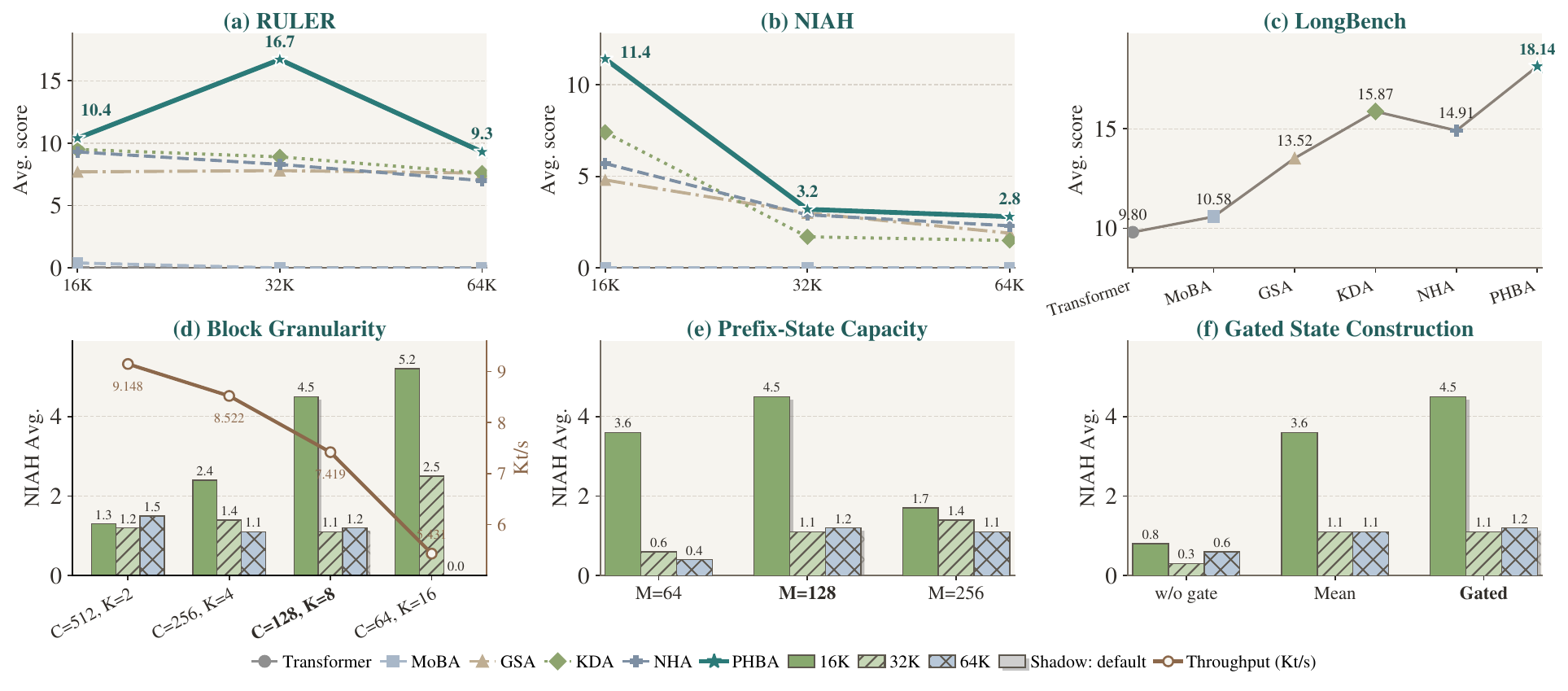}
    \caption{\small
\textbf{Long-context results and ablations.}
(a)--(b) RULER/NIAH averages at 16K/32K/64K; (c) LongBench average.
(d)--(f) Ablations of block granularity, prefix-state capacity, and gated state construction; (d) also reports throughput (1.3B, BS=1, 8K).
Bars denote 16K/32K/64K results; shadows mark default settings.
}
    \label{fig:phba_results}
\end{figure*}
% \paragraph{Long-Context Understanding.}
% On LongBench, PHBA achieves the best overall average score of 18.14 at the
% 1.3B scale.
% The improvements are particularly evident on single-document QA,
% multi-document QA, summarization, and few-shot tasks, while PHBA remains
% competitive on code benchmarks.
% This indicates that the gains of PHBA extend beyond controlled retrieval tasks
% to real-world long-context understanding.
\subsection{Ablation and Design Analysis}
\label{sec:exp-ablation}

Figure~\ref{fig:phba_results} (d)--(f) summarizes the long-context
ablations of block granularity, prefix-state capacity, and gated state
construction.
Short-context ablations are included in
Table~\ref{tab:comprehensive_merged_results_fixed}, with complete
long-context results in Appendix Table~\ref{tab:reordered_long_context_final}.

\paragraph{Block Granularity.}
Under a fixed $1/8$ exact-token budget, reducing $C$ and increasing $K$
generally improves retrieval by enabling finer-grained routing
(Figure~\ref{fig:phba_results} (d);
Table~\ref{tab:comprehensive_merged_results_fixed}).
Smaller blocks, however, require more routed blocks and reduce throughput.
We therefore use $C=128$ and $K=8$ as the default quality--efficiency trade-off.

\paragraph{Prefix-State Capacity.}
Figure~\ref{fig:phba_results} (e) and
Table~\ref{tab:comprehensive_merged_results_fixed} show that performance does
not improve monotonically with state size.
$M=128$ gives the strongest and most stable results across context lengths,
while smaller states lack capacity and larger states provide no consistent
benefit.

\paragraph{Gated State Construction.}
Figure~\ref{fig:phba_results} (f) and
Table~\ref{tab:comprehensive_merged_results_fixed} show that the learned gated
recurrence consistently outperforms ungated and mean-aggregation variants,
especially under long-context extrapolation.
This highlights the importance of adaptive retention and writing for
prefix-state construction.

\paragraph{Token--State Composition.}
MoBA retrieves distant tokens without an aligned summary of their preceding
context, whereas NHA combines recurrent memory with exact attention restricted
to a fixed local window.
PHBA instead couples routed historical blocks with their block-aligned prefix
states, yielding stronger retrieval and long-context performance in
Figure~\ref{fig:phba_results} (a)--(c) and
Table~\ref{tab:comprehensive_merged_results_fixed}.

\begin{figure*}[t]
\centering

\definecolor{transformerblue}{RGB}{31,119,180}
\definecolor{gsagray}{RGB}{105,105,105}
\definecolor{swaorange}{RGB}{230,108,32}
\definecolor{nhapurple}{RGB}{125,91,166}
\definecolor{mobagreen}{RGB}{72,135,92}
\definecolor{phbared}{RGB}{200,65,65}

\begin{tikzpicture}
\begin{axis}[
    width=0.96\textwidth,
    height=6.5cm,
    xlabel={Sequence length},
    ylabel={Training throughput (Kt/s)},
    xmin=0.9,
    xmax=4.1,
    ymin=0,
    ymax=16.2,
    xtick={1,2,3,4},
    xticklabels={2K,4K,8K,16K},
    ytick={0,2,4,6,8,10,12,14,16},
    grid=major,
    major grid style={black!10},
    axis line style={black!60},
    tick style={black!60},
    tick label style={font=\scriptsize},
    label style={font=\small},
    every axis plot/.append style={
        line width=1.35pt,
        mark size=2.5pt
    },
    legend columns=5,
    legend cell align=left,
    legend style={
        at={(0.5,-0.18)},
        anchor=north,
        font=\fontsize{6.5}{7.2}\selectfont,
        draw=black!25,
        fill=white,
        fill opacity=0.97,
        text opacity=1,
        rounded corners=1pt,
        inner xsep=4pt,
        inner ysep=2.5pt,
        row sep=1.2pt,
        column sep=8pt
    },
    clip=true
]

% Transformer
\addplot[
    color=transformerblue,
    mark=*
]
coordinates {
    (1,13.795)
    (2,12.461)
    (3,10.329)
    (4,7.796)
};
\addlegendentry{Transformer}

% SWA, fixed w=64
\addplot[
    color=swaorange,
    mark=triangle*
]
coordinates {
    (1,15.242)
    (2,15.237)
    (3,15.199)
    (4,15.217)
};
\addlegendentry{SWA}

% NHA, fixed w=64
\addplot[
    color=nhapurple,
    mark=diamond*
]
coordinates {
    (1,11.499)
    (2,11.508)
    (3,11.403)
    (4,11.552)
};
\addlegendentry{NHA}

% MoBA, fixed K=4
\addplot[
    color=mobagreen,
    mark=square*
]
coordinates {
    (1,10.238)
    (2,10.953)
    (3,10.883)
    (4,10.193)
};
\addlegendentry{MoBA ($K=4$)}

% PHBA, fixed K=4
\addplot[
    color=phbared,
    mark=star,
    very thick,
    line width=1.75pt
]
coordinates {
    (1,11.031)
    (2,9.863)
    (3,9.559)
    (4,8.790)
};
\addlegendentry{\textbf{PHBA ($K=4$)}}

% GSA
\addplot[
    color=gsagray,
    mark=o
]
coordinates {
    (1,11.395)
    (2,11.397)
    (3,11.344)
    (4,11.246)
};
\addlegendentry{GSA}

% SWA, fixed sparsity 1/8
\addplot[
    color=swaorange,
    dashed,
    mark=triangle
]
coordinates {
    (1,14.959)
    (2,14.562)
    (3,13.739)
    (4,12.455)
};
\addlegendentry{SWA ($1/8$)}

% NHA, fixed sparsity 1/8
% 0 denotes OOM
\addplot[
    color=nhapurple,
    dashed,
    mark=diamond
]
coordinates {
    (1,9.319)
    (2,0)
    (3,0)
    (4,0)
};
\addlegendentry{NHA ($1/8$)}

% MoBA, fixed C=256
\addplot[
    color=mobagreen,
    dashed,
    mark=square
]
coordinates {
    (1,13.447)
    (2,12.151)
    (3,10.805)
    (4,9.061)
};
\addlegendentry{MoBA ($C=256$)}

% PHBA, fixed C=256
\addplot[
    color=phbared,
    dashed,
    mark=star,
    very thick,
    line width=1.75pt
]
coordinates {
    (1,12.541)
    (2,11.124)
    (3,9.602)
    (4,7.987)
};
\addlegendentry{\textbf{PHBA ($C=256$)}}

\end{axis}
\end{tikzpicture}

\vspace{8pt}

\caption{
Training efficiency on a single NVIDIA H20-96GB GPU.
For MoBA and PHBA, both configurations maintain a fixed $1/8$ sparse
token budget: $K=4$ dynamically adjusts the block size $C$, whereas
$C=256$ dynamically adjusts the number of routed blocks $K$.
SWA ($1/8$) and NHA ($1/8$) similarly scale the window size with
sequence length, while SWA and NHA use a fixed window size of $64$.
All state-based models use $M=128$ memory slots.
Zero throughput for NHA ($1/8$) denotes out-of-memory (OOM).
}
\label{fig:throughput_phba}
\end{figure*}
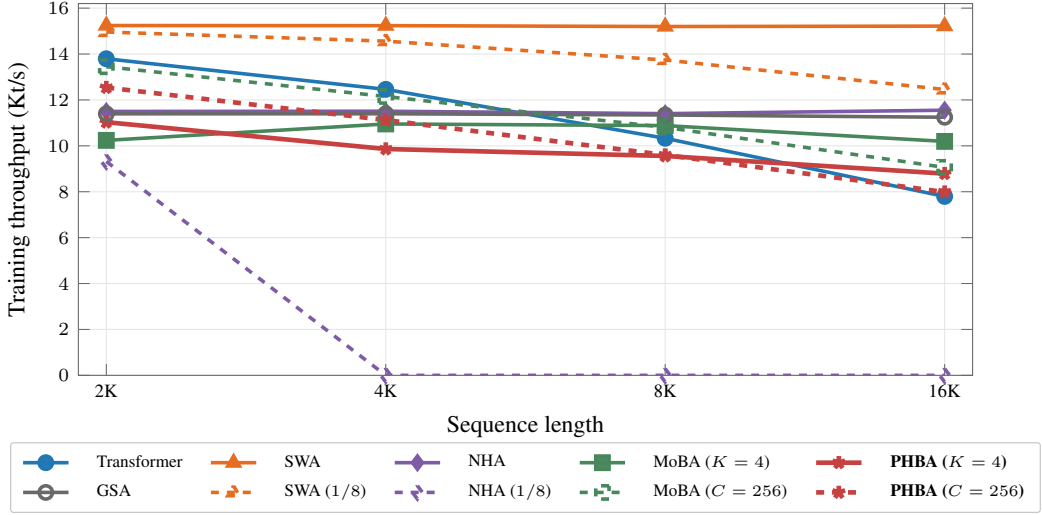

\subsection{Training Efficiency.}
Figure~\ref{fig:throughput_phba} compares end-to-end training throughput,
with detailed benchmark settings and measurement protocols provided in
Sec.~\ref{sec:e-b-p}.
With a fixed 64-token window, NHA is about $24\%$ slower than SWA.
Under the matched $1/8$ budget, PHBA is $18.4$--$34.6\%$ faster than NHA
at 2K, while NHA runs out of memory from 4K onward in the evaluated
implementation.
Relative to MoBA, adding block-aligned prefix states incurs only
$6.7$--$11.9\%$ overhead for $C=256$ across 2K--16K.
With $K=4$, PHBA reaches $8.79$ Kt/s at 16K, $12.8\%$ higher than full
attention, while GSA remains nearly length-independent due to its recurrent
computation.

\subsection{Related Work}
Efficient sequence models replace dense attention with structured or recurrent
state updates, including S4/S5, RetNet, Mamba/Mamba-2, GLA, HGRN2, GSA, and
recent gated or delta-rule variants such as DeltaNet, KDA, and Gated
DeltaNet-2~\citep{
gu2021efficiently,
smith2022simplified,
sun2023retentive,
gu2023mamba,
dao2024transformers,
yang2023gated,
qin2024hgrn2,
zhang2024gated,
yang2024parallelizing,
team2025kimi,
hatamizadeh2026gated}.
A complementary line reduces softmax-attention cost through sparse connectivity,
from Sparse Transformer, Reformer, Longformer, BigBird, and Routing Transformer
to MoBA's query-dependent block routing
~\citep{
child2019generating,
kitaev2020reformer,
beltagy2020longformer,
zaheer2020big,
roy2021efficient,
lu2026moba}.
Hybrid architectures such as BASED, Griffin, and NHA combine recurrent or
linear memory with local attention
~\citep{arora2024simple,de2024griffin,du2026native};
NHA further places recurrent key--value slots and sliding-window tokens under a
shared softmax. In this work, PHBA builds on these lines by pairing content-routed historical
blocks with block-aligned prefix states, so each selected block is accompanied
by a compressed summary of its preceding context.
\section{Conclusion}

We introduced Prefix-State Hybrid Block Attention (PHBA), which combines
content-dependent block-sparse retrieval with block-aligned recurrent prefix
states for efficient long-context modeling.
PHBA jointly attends to routed historical tokens and the compressed context
preceding them, providing both fine-grained retrieval and broader historical
information within a unified softmax.
A hardware-aware implementation further enables efficient chunkwise state
construction and sparse execution without materializing packed payloads.
Experiments demonstrate consistent improvements in retrieval and long-context
understanding while preserving competitive language modeling quality and
training efficiency.
\subsection*{AI Use Statement}

In this work, we used generative AI tools solely to aid and polish the
writing. We did not use generative AI tools for research design,
methodology, experiments, data analysis, or the generation of results.
All AI-assisted content was reviewed by the authors.
We take full responsibility for the final content of this work.

\subsection*{Ethics statement}

This work studies efficient sequence modeling architectures and does not
involve human subjects or the collection of sensitive personal data.
Experiments are conducted using publicly available training and evaluation
datasets. We are not aware of additional ethical concerns specific to the
proposed method beyond those generally associated with large language models.

\subsection*{Reproducibility Statement}

We provide detailed specifications of PHBA in the main paper, including its
routing, prefix-state construction, token--state attention, and efficient
execution. The appendix contains the chunkwise derivation, complete operator
pseudocode, training and evaluation settings, throughput benchmark protocol,
and full experimental results. Together, these materials are intended to
support faithful reproduction of the proposed method and experiments. The
source code and implementation will be made publicly available upon acceptance.

\subsubsection*{Author Contributions}
If you'd like to, you may include  a section for author contributions as is done
in many journals. This is optional and at the discretion of the authors.

\subsubsection*{Acknowledgments}
Use unnumbered third level headings for the acknowledgments. All
acknowledgments, including those to funding agencies, go at the end of the paper.

\bibliography{iclr2027_conference}
\bibliographystyle{iclr2027_conference}

\appendix
\section{Appendix}

\subsection{PHBA Operator Pseudocode}
\label{sec:appendix-phba-kernels}

This appendix expands the PHBA kernels in
Section~\ref{sec:method-chunkwise}. We write $C$ for the token block
size, $M$ for the number of prefix-state slots, and $H$ for the number of
heads. A logical block is either a token block or a prefix-state block. A
physical tile is the smaller query, memory, or feature tile loaded from HBM to
SRAM/registers by one Triton program.

The important implementation choice is that PHBA does not build packed sparse
payload tensors. The forward kernel consumes the query-centric historical
routes $\bm R^c$ directly and loads routed token and prefix-state blocks from
their original HBM layouts. During training, the same routing decisions are
reformatted into block-major integer metadata
$\bm{\rho},\mathrm{Offset},C_{\rm blk}$ for efficient sparse backward.
The generic symbols $\bm{\rho},\mathrm{Offset},C_{\rm blk}$ denote either
the historical-token metadata
$(\bm{\rho}^{\rm tok},\mathrm{Offset}^{\rm tok},C^{\rm tok}_{\rm blk})$
or the prefix-state metadata
$(\bm{\rho}^{\rm st},\mathrm{Offset}^{\rm st},C^{\rm st}_{\rm blk})$
from the main text.

\subsubsection{Chunkwise Parallel Prefix-State Memory Construction}
\label{sec:appendix-phba-state-kernel}

This subsection derives the chunkwise parallel form of the prefix-state
recurrence in Eq.~\ref{eq:phba-prefix-update}. For each token $t$, PHBA
maintains the combined slot-state matrix
$\bm{S}_t=\big[\widetilde{\bm{K}}_t,\widetilde{\bm{V}}_t\big]
\in\mathbb{R}^{M\times(d+d_v)}$ and writes
$\bm{z}_t=\big[\bm{k}^{\mathrm{raw}}_t,\bm{v}_t\big]\in\mathbb{R}^{d+d_v}$.
The recurrent update is
\begin{equation}
\bm{S}_t
=
\operatorname{Diag}(\bm{A}_t)\bm{S}_{t-1}
+
\bm{I}_t\bm{z}_t^\top,
\label{eq:app-phba-state-recurrent}
\end{equation}
where $\bm{A}_t\in(0,1)^M$ is the retention gate and
$\bm{I}_t=\bm{1}-\bm{A}_t$ is the coupled write gate. PHBA separates memory
formation from sparse readout: the recurrence constructs boundary prefix
key/value memories, and the attention layer later selects among them.

We now derive the chunkwise parallel form. For packed inputs, the recurrence is reset at each sequence boundary
specified by $\mathrm{cu}$, so the first block of every sequence receives
a zero incoming state. For any token-wise tensor
$\bm{X}$, let $\bm{X}_{[n]}$ denote its rows in chunk $n$, where
$C_n=|\mathcal{B}_n|$. Thus
$\bm{K}^{\mathrm{raw}}_{[n]}\in\mathbb{R}^{C_n\times d}$,
$\bm{V}_{[n]}\in\mathbb{R}^{C_n\times d_v}$, and
$\bm{A}_{[n]},\bm{I}_{[n]}\in\mathbb{R}^{C_n\times M}$ stack the raw keys,
values, retention gates, and write gates in this chunk. Define
$\bm{Z}_{[n]}=\big[\bm{K}^{\mathrm{raw}}_{[n]},\bm{V}_{[n]}\big]
\in\mathbb{R}^{C_n\times(d+d_v)}$.
Partially unrolling Eq.~\ref{eq:app-phba-state-recurrent} for the $r$-th row
inside chunk $n$ gives
\begin{equation}
\bm{S}_{[n],r}
=
\operatorname{Diag}\!\left(
\prod_{i=1}^{r}\bm{A}_{[n],i}
\right)
\bm{S}_{[n],0}
+
\sum_{i=1}^{r}
\left(
\bm{I}_{[n],i}\odot
\prod_{j=i+1}^{r}\bm{A}_{[n],j}
\right)
\bm{z}_{[n],i}^{\top},
\label{eq:app-phba-state-prefix}
\end{equation}
where
$\bm{z}_{[n],i}
=
\big[\bm{k}_{[n],i}^{\mathrm{raw}},\bm{v}_{[n],i}\big]$.
Here $\bm{S}_{[n],0}$ is the boundary state entering chunk $n$, and
$\bm{S}_{[n],C_n}$ is the boundary state leaving it.
For brevity, we write
$\bm{S}_{[n]}\equiv\bm{S}_{[n],0}$ and
$\bm{S}_{[n+1]}\equiv\bm{S}_{[n],C_n}$.

As in chunkwise GLA, define the cumulative retention products
\begin{align}
\overrightarrow{\bm{A}}_{[n],r}
&=
\prod_{j=1}^{r}\bm{A}_{[n],j},
&
\overleftarrow{\bm{A}}_{[n],r}
&=
\prod_{j=r+1}^{C_n}\bm{A}_{[n],j}.
\label{eq:app-phba-state-gates}
\end{align}
All products over the gate vectors are elementwise.
The forward product $\overrightarrow{\bm{A}}$ propagates boundary memory across
the chunk, while the backward product $\overleftarrow{\bm{A}}$ decays each
within-chunk write to the chunk boundary. Let
$\overleftarrow{\bm{A}}_{[n]}\in\mathbb{R}^{C_n\times M}$ stack these backward
products. The compact write matrix is
\begin{equation}
\bm{W}_{[n]}
=
\bm{I}_{[n]}\odot\overleftarrow{\bm{A}}_{[n]}
\in\mathbb{R}^{C_n\times M},
\qquad
\bm{\Delta}_{[n]}
=
\bm{W}_{[n]}^\top\bm{Z}_{[n]}
\in\mathbb{R}^{M\times(d+d_v)}.
\label{eq:app-phba-state-delta}
\end{equation}
Then the chunkwise boundary update becomes
\begin{equation}
\bm{S}_{[n+1]}
=
\underbrace{
\operatorname{Diag}\!\left(\overrightarrow{\bm{A}}_{[n],C_n}\right)
\bm{S}_{[n]}}_{\text{inter-chunk recurrence}}
+
\underbrace{\bm{\Delta}_{[n]}}_{\text{intra-chunk parallel write}} .
\label{eq:app-phba-state-boundary}
\end{equation}
Equation~\ref{eq:app-phba-state-boundary} is the form used for parallel
training. Only the boundary state $\bm{S}_{[n]}$ is sequential across chunks.
Inside each chunk, the writes reduce to matrix products over chunk rows and
state slots. Thus PHBA materializes one prefix key/value state per chunk
boundary rather than one recurrent state per token.
For numerical stability, the implementation forms these cumulative retention
factors from chunk-local prefix sums of $\log\bm{A}$ rather than explicitly
multiplying gate values.

\subsubsection{Routing Metadata Kernels}
\label{sec:appendix-phba-metadata}

PHBA uses a three-stage routing pipeline: compute block centroids,
select top-$K_h$ historical block ids with a tiled score kernel, and
reformat the query-centric routes into block-major metadata for sparse
backward. The dense query--centroid score matrix is never materialized
to HBM. The main routing procedure also constructs
$\mathrm{PrefixValid}\in\{0,1\}^{N_c}$, where
$\mathrm{PrefixValid}[b]=1$ iff block $b$ has a nonempty causal prefix.

\paragraph{Block Centroids.}
Algorithm~\ref{alg:phba-app-centroid} computes one routing centroid for
each block and attention head. Blocks are formed independently within
each packed sequence specified by $\mathrm{cu}$, and only valid rows
contribute to the centroid.

\begin{algorithm}[h]
\caption{Fused Key-Block Centroid Computation}
\label{alg:phba-app-centroid}
\begin{algorithmic}[1]

\Require RoPE keys
$\bm{K}^{\mathrm{rope}}\in\mathbb{R}^{T\times H\times d}$
in HBM.
Block-row pointers
$\mathrm{BlockPtr}\in\mathbb{Z}^{N_c+1}$ induced by
$\mathrm{cu}$ and block size $C$.

\State Allocate centroid tensor
$\bar{\bm{K}}^{\mathrm{rope}}
\in\mathbb{R}^{N_c\times H\times d}$ in HBM.

\For{each block $j$ and head $h$ \textbf{in parallel}}
    \State Set
    $\mathcal{B}_j=
    [\mathrm{BlockPtr}[j],\mathrm{BlockPtr}[j+1])$.
    \State Stream valid rows of
    $\bm{K}^{\mathrm{rope}}_{\mathcal{B}_j,h}$
    from HBM through SRAM and accumulate their sum in fp32 registers.
    \State Divide by $|\mathcal{B}_j|$ on chip and write
    $\bar{\bm{K}}^{\mathrm{rope}}_{j,h}$ to HBM.
\EndFor

\State \Return $\bar{\bm{K}}^{\mathrm{rope}}$.

\end{algorithmic}
\end{algorithm}

\paragraph{Historical Block Selection.}
Algorithm~\ref{alg:phba-app-topk} scans centroid tiles and maintains a
running top-$K_h$ list for each query and head. Only blocks from the same
packed sequence and strictly preceding the query's current block are
eligible. If fewer than $K_h$ blocks are available, unused route entries
remain $-1$. The discrete route indices are treated as non-differentiable
metadata.

\begin{algorithm}[h]
\caption{Tiled Top-$K$ Historical Block Selection}
\label{alg:phba-app-topk}
\begin{algorithmic}[1]

\Require RoPE queries
$\bm{Q}^{\mathrm{rope}}\in\mathbb{R}^{T\times H\times d}$
and centroids
$\bar{\bm{K}}^{\mathrm{rope}}
\in\mathbb{R}^{N_c\times H\times d}$ in HBM.
Boundaries $\mathrm{cu}$.
Block-row pointers $\mathrm{BlockPtr}$.
Current block ids $\bm{c}$.
Total exact-block budget $K$.
Tile sizes $B_r,B_c$.

\State Set historical routing budget $K_h\leftarrow K-1$.

\State For each head and packed sequence, partition query rows and
centroid rows into tiles of sizes $B_r$ and $B_c$, respectively.

\State Initialize
$\bm{R}^{c}\in\mathbb{Z}^{T\times H\times K_h}$
to $-1$ in HBM.

\For{each query tile $i$ and head $h$ \textbf{in parallel}}
    \State Load $\bm{Q}^{\mathrm{rope}}_{i,h}$ from HBM to SRAM and
    initialize top-$K_h$ scores to $-\infty$ and block ids to $-1$.

    \For{each centroid tile $j$ in the same packed sequence}
        \State Load
        $\bar{\bm{K}}^{\mathrm{rope}}_{j,h}$ from HBM to SRAM.
        \State Compute
        $\bm{S}_{ij,h}\leftarrow
        \bm{Q}^{\mathrm{rope}}_{i,h}
        (\bar{\bm{K}}^{\mathrm{rope}}_{j,h})^\top$
        on chip.
        \State Apply row-wise $b<c(t)$ and valid-row masks and update
        the running top-$K_h$ distinct block ids in registers.
    \EndFor

    \State Write the final block ids to
    $\bm{R}^{c}$ in HBM.
\EndFor

\State \Return $\bm{R}^{c}$.

\end{algorithmic}
\end{algorithm}

\paragraph{Block-Major Metadata.}
Algorithm~\ref{alg:phba-app-sparse-metadata} inverts the query-centric
routing relation. It counts query--head rows assigned to each logical
block--head pair, prefix-sums the counts to obtain contiguous ranges,
and scatters flattened query--head row ids into these ranges.
Only integer metadata is materialized; token and state payloads remain
in their original HBM layouts.

\begin{algorithm}[h]
\caption{Reformat Routes to Block-Major Varlen Metadata}
\label{alg:phba-app-sparse-metadata}
\begin{algorithmic}[1]

\Require Query-centric routes
$\bm{R}\in\mathbb{Z}^{T\times H\times K_{\rm sel}}$
in HBM, with distinct valid block ids within each query--head row.
Number of logical blocks $N_c$.
Number of heads $H$.

\State Initialize
$C_{\rm blk}\in\mathbb{Z}^{N_cH}$ to zero in HBM.

\For{each route slot $(t,h,k)$ \textbf{in parallel}}
    \State Set $b\leftarrow\bm{R}[t,h,k]$.
    \If{$b\ge0$}
        \State $\mathrm{atomicAdd}(\&C_{\rm blk}[bH+h],1)$.
    \EndIf
\EndFor

\State Exclusive prefix-sum $C_{\rm blk}$ to obtain
$\mathrm{Offset}\in\mathbb{Z}^{N_cH+1}$ with
$\mathrm{Offset}[0]=0$.

\State Set $N_{\rm edge}\leftarrow\mathrm{Offset}[N_cH]$,
allocate
$\bm{\rho}\in\mathbb{Z}^{N_{\rm edge}}$,
and initialize
$\mathrm{Cursor}\in\mathbb{Z}^{N_cH}$ to zero in HBM.

\For{each route slot $(t,h,k)$ \textbf{in parallel}}
    \State Set $b\leftarrow\bm{R}[t,h,k]$.
    \If{$b\ge0$}
        \State Set $e\leftarrow bH+h$, $r\leftarrow tH+h$, and
        $p\leftarrow\mathrm{atomicAdd}(\&\mathrm{Cursor}[e],1)$.
        \State Store
        $r$ into $\bm{\rho}[\mathrm{Offset}[e]+p]$.
    \EndIf
\EndFor

\State \Return
$\bm{\rho},\mathrm{Offset},C_{\rm blk}$.

\end{algorithmic}
\end{algorithm}

\subsubsection{Query-Tile Fused Sparse Forward Kernel}
\label{sec:appendix-phba-sparse-forward}

Algorithm~\ref{alg:phba-app-sparse-forward} implements the PHBA forward
pass directly from the query-centric routes $\bm R^c$.
Each query tile starts from the block-local causal FlashAttention output
and log-sum-exp, then extends the same online-softmax accumulators with
the current-block prefix state, routed historical token blocks, and their
aligned prefix states.

Within a query tile, the $B_rK_h$ historical route entries use a
fixed-capacity schedule rather than dynamic compaction. A leader mask
marks the first valid occurrence of each historical block id, allowing
one loaded block to serve all rows in the tile that route to it.
The token branch uses
$(\bm Q^{\mathrm{rope}},\bm K^{\mathrm{rope}},\bm V)$,
whereas the state branch uses
$(\bm Q^{\mathrm{raw}},\bm P^K,\bm P^V)$.
Payload tensors remain in their original HBM layouts.

\begin{algorithm}[h]
\caption{Query-Tile Fused Token--State Sparse Forward}
\label{alg:phba-app-sparse-forward}
\begin{algorithmic}[1]

\Require
$\bm{Q}^{\mathrm{raw}},\bm{Q}^{\mathrm{rope}}
\in\mathbb{R}^{T\times H\times d}$,
$\bm{K}^{\mathrm{rope}}\in\mathbb{R}^{T\times H\times d}$,
and
$\bm{V}\in\mathbb{R}^{T\times H\times d_v}$ in HBM.
Prefix states $\bm{P}^{K},\bm{P}^{V}$ in HBM.
Routes $\bm{R}^{c}$ and current block ids $\bm{c}$.
Block-row pointers $\mathrm{BlockPtr}$.
Prefix-validity mask $\mathrm{PrefixValid}$.
Seed $\bm{O}^{\rm loc},L^{\rm loc}$.
Tile sizes $B_r,B_c$.

\State Divide query rows within each current block and head
into contiguous tiles of size $B_r$.

\For{each query tile $\mathcal{Q}_i$ \textbf{in parallel}}

    \State Let $c_i$ and $h_i$ be the current block and head
    shared by rows in $\mathcal{Q}_i$.

    \State Load
    $\bm{Q}^{\mathrm{rope}}_i,\bm{Q}^{\mathrm{raw}}_i$
    and
    $\bm{O}^{\rm loc}_i,L^{\rm loc}_i$
    from HBM.

    \State Initialize row-wise fp32 online-softmax accumulators as
    $m_i\leftarrow L^{\rm loc}_i$,
    $\ell_i\leftarrow\bm 1$, and
    $\bm U_i\leftarrow\bm O^{\rm loc}_i$.
    \Statex \Comment{$e^{L_i^{\rm loc}}\bm O_i^{\rm loc}$ equals the
    unnormalized local-attention numerator.}

    \State Load historical routes
    $\bm{R}^{c}_i$ to SRAM, setting padded query rows to $-1$.

    \State Set
    $\mu_i^{\rm cur}\leftarrow\mathrm{PrefixValid}[c_i]$
    and divide the $M$ current-prefix slots into tiles of size $B_c$.

    \For{each current-prefix-state tile}
        \State Load the corresponding
        $\bm{P}^{K}_{c_i,h_i},\bm{P}^{V}_{c_i,h_i}$
        rows from HBM to SRAM under validity and tail masks.
        \State Compute scores using
        $\bm{Q}^{\mathrm{raw}}_i/\sqrt d$,
        mask invalid scores to $-\infty$, and update
        $(m_i,\ell_i,\bm U_i)$ on chip.
    \EndFor

    \State Flatten $\bm{R}^{c}_i$ into
    $\bm b_i=(b_{i,1},\ldots,b_{i,J})$ with $J=B_rK_h$.

    \State Compute leader mask
    $\lambda_{i,p}\leftarrow
    \mathbbm{1}[b_{i,p}\ge0]\,
    \mathbbm{1}[b_{i,p}\notin\{b_{i,1},\ldots,b_{i,p-1}\}]$
    on chip.

    \For{each fixed candidate slot $p=1,\ldots,J$}

        \State Set
        $b\leftarrow b_{i,p}$,
        $\tilde b\leftarrow\max(b,0)$, and
        $\bm\chi_{i,p}[r]\leftarrow
        \lambda_{i,p}\,
        \mathbbm{1}[\exists k:\bm R^c_i[r,k]=b]$.

        \State Set
        $\mathcal{B}_{\tilde b}\leftarrow
        [\mathrm{BlockPtr}[\tilde b],
        \mathrm{BlockPtr}[\tilde b+1])$
        and divide it into token-memory tiles of size $B_c$.

        \For{each token-memory tile}
            \State Load the corresponding
            $\bm{K}^{\mathrm{rope}}_{\mathcal{B}_{\tilde b},h_i},
            \bm{V}_{\mathcal{B}_{\tilde b},h_i}$
            rows from HBM to SRAM under leader and tail masks.
            \State Compute scores using
            $\bm{Q}^{\mathrm{rope}}_i/\sqrt d$,
            mask rows with $\bm\chi_{i,p}[r]=0$ and invalid tails,
            and update $(m_i,\ell_i,\bm U_i)$ on chip.
        \EndFor

        \State Set
        $\mu_{i,p}^{\rm st}\leftarrow
        \lambda_{i,p}\mathrm{PrefixValid}[\tilde b]$
        and divide the $M$ aligned-prefix slots into tiles of size $B_c$.

        \For{each prefix-state tile}
            \State Load the corresponding
            $\bm{P}^{K}_{\tilde b,h_i},
            \bm{P}^{V}_{\tilde b,h_i}$
            rows from HBM to SRAM under validity and tail masks.
            \State Compute scores using
            $\bm{Q}^{\mathrm{raw}}_i/\sqrt d$,
            mask rows with $\bm\chi_{i,p}[r]=0$ or
            $\mu_{i,p}^{\rm st}=0$, and update
            $(m_i,\ell_i,\bm U_i)$ on chip.
        \EndFor

    \EndFor

    \State Finalize
    $\bm O_i\leftarrow\bm U_i/\ell_i$ and
    $L_i\leftarrow m_i+\log\ell_i$,
    and write $\bm O_i,L_i$ to HBM.

\EndFor

\State \Return output $\bm O$ and log-sum-exp $L$.

\end{algorithmic}
\end{algorithm}

\subsubsection{Sparse Backward Kernel}
\label{sec:appendix-phba-backward}

The backward pass complements the query-centric forward with a
block-major schedule. Algorithm~\ref{alg:phba-app-backward} uses
$\bm{\rho}$ and $\mathrm{Offset}$ to enumerate the query--head rows
accessing each token or prefix-state block. Sparse scores and
probabilities are recomputed on chip from the final joint log-sum-exp
$L$, avoiding materialized sparse probability tensors.

For compactness, let
$(\bm Q^{\rm tok},\bm K^{\rm tok}_{j,h},\bm V^{\rm tok}_{j,h})$
denote
$(\bm Q^{\rm rope},
\bm K^{\rm rope}_{\mathcal B_j,h},
\bm V_{\mathcal B_j,h})$,
and let
$(\bm Q^{\rm st},\bm K^{\rm st}_{j,h},\bm V^{\rm st}_{j,h})$
denote
$(\bm Q^{\rm raw},\bm P^K_{j,h},\bm P^V_{j,h})$,
where
$\mathcal B_j=
[\mathrm{BlockPtr}[j],\mathrm{BlockPtr}[j+1])$.
The two logical memories therefore contain
$|\mathcal B_j|$ token rows and $M$ prefix-state slots, respectively.

\begin{algorithm}[h]
\caption{Metadata-Driven Sparse Backward}
\label{alg:phba-app-backward}
\begin{algorithmic}[1]

\Require Token tensors
$\bm{Q}^{\mathrm{raw}},\bm{Q}^{\mathrm{rope}},
\bm{K}^{\mathrm{raw}},\bm{K}^{\mathrm{rope}},\bm V$
in HBM.
Prefix states $\bm P^K,\bm P^V$ and retention gates $\bm A$.
Block-row pointers $\mathrm{BlockPtr}$ and boundaries $\mathrm{cu}$.
Final output and gradient
$\bm O,\bm{dO}\in\mathbb{R}^{T\times H\times d_v}$,
joint log-sum-exp $L\in\mathbb{R}^{T\times H}$,
block-major metadata
$(\bm\rho^\beta,\mathrm{Offset}^\beta,C_{\rm blk}^\beta)$
for $\beta\in\{\mathrm{tok},\mathrm{st}\}$,
and tile sizes $B_r,B_c$.

\State Compute
$D\leftarrow\mathrm{rowsum}(\bm{dO}\circ\bm O)$
and initialize query and memory-gradient buffers to zero.

\State Run block-local FlashAttention-style backward using the final
joint $L,D$ and accumulate into the token-gradient buffers.

\For{$\beta\in\{\mathrm{tok},\mathrm{st}\}$}

    \For{each logical block $j$ and head $h$ with
    $C_{\rm blk}^{\beta}[jH+h]>0$ \textbf{in parallel}}

        \State Read routed query rows
        $\mathcal R^\beta_{j,h}$ from
        $\bm\rho^\beta,\mathrm{Offset}^\beta$
        and partition
        $(\bm K^\beta_{j,h},\bm V^\beta_{j,h})$
        into memory tiles of size $B_c$.

        \For{each memory tile $\mathcal M$}

            \State Load
            $\bm K^\beta_{\mathcal M,h},
            \bm V^\beta_{\mathcal M,h}$
            and initialize
            $\bm{dK}^\beta_{\mathcal M,h},
            \bm{dV}^\beta_{\mathcal M,h}\leftarrow0$.

            \For{each $B_r$-sized query tile in
            $\mathcal R^\beta_{j,h}$}

                \State Gather
                $\bm Q^\beta,\bm{dO},L,D$ to SRAM.

                \State Recompute
                $\bm S\leftarrow
                \bm Q^\beta
                (\bm K^\beta_{\mathcal M,h})^\top/\sqrt d$
                and mask invalid tail columns.

                \State Compute
                $\bm\Pi\leftarrow\exp(\bm S-L)$,
                $\bm{d\Pi}\leftarrow
                \bm{dO}(\bm V^\beta_{\mathcal M,h})^\top$,
                and
                $\bm{dS}\leftarrow
                \bm\Pi\circ(\bm{d\Pi}-D)$.

                \State Accumulate
                $\bm{dV}^\beta_{\mathcal M,h}
                \mathrel{+}=\bm\Pi^\top\bm{dO}$
                and
                $\bm{dK}^\beta_{\mathcal M,h}
                \mathrel{+}=\bm{dS}^\top\bm Q^\beta/\sqrt d$.

                \State Atomically scatter-add
                $\bm{dS}\bm K^\beta_{\mathcal M,h}/\sqrt d$
                to the corresponding $\bm{dQ}^\beta$ rows.

            \EndFor

            \State Accumulate
            $\bm{dK}^\beta_{\mathcal M,h},
            \bm{dV}^\beta_{\mathcal M,h}$
            into HBM.

        \EndFor
    \EndFor
\EndFor

\State Stack direct prefix-state adjoints as
$\bm{dP}_{[n]}\leftarrow
[\bm{dP}^{K}_{[n]},\bm{dP}^{V}_{[n]}]$.

\State Traverse blocks in reverse within each packed sequence,
with zero propagated adjoint at the terminal boundary.

\For{each block $[n]$ in reverse order}

    \State Set
    $\bar{\bm S}_{[n+1]}
    \leftarrow
    \bm{dP}_{[n+1]}+
    \bar{\bm S}^{\,\mathrm{prop}}_{[n+1]}$,
    with $\bm{dP}_{[n+1]}=0$ at an unexposed terminal boundary.

    \State Load the incoming boundary state
    $\bm S_{[n]}\leftarrow
    [\bm P^K_{[n]},\bm P^V_{[n]}]$.

    \State Recompute the chunk-local cumulative retention factors and
    $\bm W_{[n]}\leftarrow
    \bm I_{[n]}\odot\overleftarrow{\bm A}_{[n]}$
    from the retained gate values.

    \State Backpropagate
    $\bar{\bm S}_{[n+1]}$ once through
    $\bm S_{[n+1]}\leftarrow
    \operatorname{Diag}(\overrightarrow{\bm A}_{[n],C_n})
    \bm S_{[n]}+\bm W_{[n]}^\top\bm Z_{[n]}$,
    obtaining
    $\bar{\bm S}^{\,\mathrm{prop}}_{[n]}
    \leftarrow
    \operatorname{Diag}(\overrightarrow{\bm A}_{[n],C_n})
    \bar{\bm S}_{[n+1]}$
    and gradients to
    $\bm Z_{[n]}=[\bm K^{\mathrm{raw}}_{[n]},\bm V_{[n]}]$
    and $\bm A_{[n]}$, including the coupled write dependence
    $\bm I_{[n]}=\bm 1-\bm A_{[n]}$.

\EndFor

\State Backpropagate the retention-gate gradients through
$\bm A=\sigma(\bm a)^{1/\tau}$.

\State Backpropagate token-path query/key gradients through RoPE,
combine them with the raw query/key gradients from the state path,
and combine token-attention and state-construction gradients for $\bm V$.

\State \Return accumulated token, state, and gate gradients.

\end{algorithmic}
\end{algorithm}

\begin{table*}[t]
    \centering

    \caption{
        Long-context evaluation on RULER, NIAH, and LongBench.
        Metrics are multiplied by 100.
        \textbf{Bold} and \underline{underline} indicate the best and
        second-best results within each model scale, respectively.
        PHBA (origin) uses $M=128$, $C=128$, and $K=8$.
    }

    \vspace{-0.35em}

    \begin{small}

    % ============================================================
    % Compact global spacing
    % ============================================================
    \renewcommand{\multirowsetup}{\centering}
    \setlength{\aboverulesep}{0.25ex}
    \setlength{\belowrulesep}{0.25ex}

    % ============================================================
    % PART I: RULER
    % ============================================================
    \renewcommand{\arraystretch}{0.98}
    \setlength{\tabcolsep}{3.0pt}

    \begin{adjustbox}{width=1.0\textwidth,center}
    \begin{tabular}{
        @{}
        l|
        ccc|
        ccc|
        ccc|
        ccc
        @{}
    }

        \toprule

        \multicolumn{13}{c}{
            \textbf{PART I: RULER Tasks (Long Context)}
        }
        \\

        \midrule

        \multirow{2}{*}{\textbf{Model \& Scale}}
        & \multicolumn{3}{c|}{\textbf{CWE}}
        & \multicolumn{3}{c|}{\textbf{HotpotQA}}
        & \multicolumn{3}{c|}{\textbf{SQuAD}}
        & \multicolumn{3}{c}{\textbf{Avg}}
        \\

        \cmidrule(lr){2-4}
        \cmidrule(lr){5-7}
        \cmidrule(lr){8-10}
        \cmidrule(lr){11-13}

        &
        16k$_\uparrow$
        & 32k$_\uparrow$
        & 64k$_\uparrow$

        & 16k$_\uparrow$
        & 32k$_\uparrow$
        & 64k$_\uparrow$

        & 16k$_\uparrow$
        & 32k$_\uparrow$
        & 64k$_\uparrow$

        & 16k$_\uparrow$
        & 32k$_\uparrow$
        & 64k$_\uparrow$
        \\

        \midrule

        \multicolumn{13}{l}{
            \textit{760M params with 50B training tokens}
        }
        \\

        \rowcolor{blue!10}
        \hspace{2mm} PHBA (origin)
        & \underline{0.1}
        & 0.0
        & \underline{0.0}
        & \textbf{12.8}
        & 10.0
        & 6.4
        & 8.8
        & \underline{6.9}
        & \textbf{5.5}
        & \cellcolor{red!7}\underline{7.2}
        & \cellcolor{red!7}\underline{5.6}
        & \cellcolor{red!7}4.0
        \\

        \hspace{4mm} w. $C=512, K=2$
        & \underline{0.1}
        & 0.0
        & \underline{0.0}
        & 11.8
        & 10.4
        & \textbf{10.4}
        & 8.0
        & 6.8
        & \textbf{5.5}
        & \cellcolor{red!7}6.6
        & \cellcolor{red!7}\textbf{5.7}
        & \cellcolor{red!7}\textbf{5.3}
        \\

        \hspace{4mm} w. $C=256, K=4$
        & \underline{0.1}
        & 0.0
        & \textbf{0.1}
        & 9.4
        & 10.6
        & \underline{9.8}
        & 4.3
        & 4.7
        & 4.3
        & \cellcolor{red!7}4.6
        & \cellcolor{red!7}5.1
        & \cellcolor{red!7}4.7
        \\

        \hspace{4mm} w. $C=64, K=16$
        & \underline{0.1}
        & 0.0
        & \underline{0.0}
        & \textbf{12.8}
        & 8.8
        & 0.4
        & \textbf{9.1}
        & \textbf{7.1}
        & 0.2
        & \cellcolor{red!7}\textbf{7.3}
        & \cellcolor{red!7}5.3
        & \cellcolor{red!7}0.2
        \\

        \hspace{4mm} w. $M=64$
        & \textbf{0.2}
        & 0.0
        & \underline{0.0}
        & 10.8
        & \textbf{11.4}
        & 9.0
        & \underline{8.9}
        & 5.2
        & 5.1
        & \cellcolor{red!7}6.6
        & \cellcolor{red!7}5.5
        & \cellcolor{red!7}4.7
        \\

        \hspace{4mm} w. $M=256$
        & \underline{0.1}
        & 0.0
        & \underline{0.0}
        & \underline{12.2}
        & 9.6
        & 9.6
        & \textbf{9.1}
        & 6.7
        & 4.9
        & \cellcolor{red!7}7.1
        & \cellcolor{red!7}5.4
        & \cellcolor{red!7}\underline{4.8}
        \\

        \hspace{4mm} w/o. gate
        & \underline{0.1}
        & 0.0
        & \underline{0.0}
        & 8.2
        & 4.4
        & 3.2
        & 4.6
        & 3.4
        & 1.9
        & \cellcolor{red!7}4.3
        & \cellcolor{red!7}2.6
        & \cellcolor{red!7}1.7
        \\

        \hspace{4mm} w. mean
        & \textbf{0.2}
        & 0.0
        & \underline{0.0}
        & 10.4
        & \underline{11.0}
        & 9.2
        & 7.0
        & 5.9
        & 4.2
        & \cellcolor{red!7}5.9
        & \cellcolor{red!7}\underline{5.6}
        & \cellcolor{red!7}4.5
        \\

        \midrule

        \multicolumn{13}{l}{
            \textit{1.3B params with 100B training tokens}
        }
        \\

        \rowcolor{gray!30}
        \hspace{2mm} Transformer
        & 0.0
        & 0.0
        & 0.0
        & 0.0
        & 0.0
        & 0.0
        & 0.0
        & 0.0
        & 0.0
        & \cellcolor{red!7}0.0
        & \cellcolor{red!7}0.0
        & \cellcolor{red!7}0.0
        \\

        \hspace{2mm} MoBA
        & 0.0
        & 0.0
        & 0.0
        & 0.0
        & 0.0
        & 0.0
        & 1.1
        & 0.1
        & 0.0
        & \cellcolor{red!7}0.4
        & \cellcolor{red!7}0.0
        & \cellcolor{red!7}0.0
        \\

        \hspace{2mm} GSA
        & 0.2
        & 0.0
        & 0.0
        & 14.0
        & 14.8
        & \textbf{13.6}
        & 9.0
        & 8.6
        & 9.2
        & \cellcolor{red!7}7.7
        & \cellcolor{red!7}7.8
        & \cellcolor{red!7}\underline{7.6}
        \\

        \hspace{2mm} KDA
        & 0.2
        & 0.0
        & \underline{0.2}
        & \underline{15.4}
        & \underline{15.0}
        & 13.0
        & \textbf{12.9}
        & \textbf{11.7}
        & \underline{9.5}
        & \cellcolor{red!7}\underline{9.5}
        & \cellcolor{red!7}\underline{8.9}
        & \cellcolor{red!7}\underline{7.6}
        \\

        \hspace{2mm} NHA
        & \underline{0.5}
        & \underline{0.1}
        & 0.0
        & \textbf{16.0}
        & \textbf{15.6}
        & \underline{13.4}
        & 11.3
        & 9.2
        & 7.7
        & \cellcolor{red!7}9.3
        & \cellcolor{red!7}8.3
        & \cellcolor{red!7}7.0
        \\

        \rowcolor{blue!10}
        \hspace{2mm} PHBA
        & \textbf{5.0}
        & \textbf{26.0}
        & \textbf{4.0}
        & 14.0
        & 13.0
        & \underline{13.4}
        & \underline{12.2}
        & \underline{11.0}
        & \textbf{10.4}
        & \cellcolor{red!7}\textbf{10.4}
        & \cellcolor{red!7}\textbf{16.7}
        & \cellcolor{red!7}\textbf{9.3}
        \\

        \bottomrule

    \end{tabular}
    \end{adjustbox}

    % ============================================================
    % PART II: NIAH
    % ============================================================
    \vspace{0.05em}

    \renewcommand{\arraystretch}{0.96}
    \setlength{\tabcolsep}{1.8pt}

    \begin{adjustbox}{width=1.0\textwidth,center}
    \begin{tabular}{
        @{}
        l|
        ccccccccc|
        cccccc|
        ccc
        @{}
    }

        \toprule

        \multicolumn{19}{c}{
            \textbf{PART II: NIAH Tasks (Long Context)}
        }
        \\

        \midrule

        \multirow{3}{*}{\textbf{Model \& Scale}}
        & \multicolumn{9}{c|}{\textbf{Multi-Key NIAH}}
        & \multicolumn{6}{c|}{\textbf{NIAH (Extra)}}
        & \multicolumn{3}{c}{\textbf{Avg}}
        \\

        \cmidrule(lr){2-10}
        \cmidrule(lr){11-16}
        \cmidrule(lr){17-19}

        &
        \multicolumn{3}{c}{\textbf{Task 1}}
        & \multicolumn{3}{c}{\textbf{Task 2}}
        & \multicolumn{3}{c|}{\textbf{Task 3}}
        & \multicolumn{3}{c}{\textbf{MultiQuery}}
        & \multicolumn{3}{c|}{\textbf{MultiValue}}
        & \multicolumn{3}{c}{\textbf{Mean}}
        \\

        &
        16k & 32k & 64k
        & 16k & 32k & 64k
        & 16k & 32k & 64k
        & 16k & 32k & 64k
        & 16k & 32k & 64k
        & 16k & 32k & 64k
        \\

        \midrule

        \multicolumn{19}{l}{
            \textit{760M params with 50B training tokens}
        }
        \\

        \rowcolor{blue!10}
        \hspace{2mm} PHBA (origin)
        & \underline{10.0} & 2.2 & \textbf{3.2}
        & 0.0 & 0.0 & 0.0
        & 0.2 & 0.0 & 0.0
        & \underline{4.8} & 1.2 & 0.3
        & \underline{7.4} & 1.9 & \textbf{2.4}
        & \cellcolor{red!7}\underline{4.5}
        & \cellcolor{red!7}1.1
        & \cellcolor{red!7}\underline{1.2}
        \\

        \hspace{4mm} w. $C=512, K=2$
        & 3.2 & 2.2 & \textbf{3.2}
        & 0.2 & 0.0 & 0.0
        & \textbf{0.4} & 0.0 & 0.0
        & 1.3 & 1.9 & \textbf{2.6}
        & 1.6 & 1.7 & 1.8
        & \cellcolor{red!7}1.3
        & \cellcolor{red!7}\underline{1.2}
        & \cellcolor{red!7}\textbf{1.5}
        \\

        \hspace{4mm} w. $C=256, K=4$
        & 3.6 & 2.2 & \textbf{3.2}
        & 0.0 & 0.0 & 0.0
        & 0.0 & 0.0 & 0.0
        & 2.9 & \underline{2.3} & 0.0
        & 5.4 & \underline{2.6} & \textbf{2.4}
        & \cellcolor{red!7}2.4
        & \cellcolor{red!7}1.4
        & \cellcolor{red!7}1.1
        \\

        \hspace{4mm} w. $C=64, K=16$
        & \textbf{10.6} & \textbf{5.8} & 0.0
        & 0.2 & 0.0 & 0.0
        & 0.0 & 0.0 & 0.0
        & \textbf{6.1} & \textbf{2.4} & 0.0
        & \textbf{9.2} & \textbf{4.3} & 0.0
        & \cellcolor{red!7}\textbf{5.2}
        & \cellcolor{red!7}\textbf{2.5}
        & \cellcolor{red!7}0.0
        \\

        \hspace{4mm} w. $M=64$
        & 7.0 & 2.2 & 1.8
        & \textbf{0.8} & 0.0 & 0.0
        & 0.0 & 0.0 & 0.0
        & 3.9 & 0.4 & 0.0
        & 6.1 & 0.2 & 0.4
        & \cellcolor{red!7}3.6
        & \cellcolor{red!7}0.6
        & \cellcolor{red!7}0.4
        \\

        \hspace{4mm} w. $M=256$
        & 3.2 & 2.2 & \textbf{3.2}
        & 0.0 & 0.0 & 0.0
        & \textbf{0.4} & 0.0 & 0.0
        & 2.4 & \underline{2.3} & 0.0
        & 2.3 & \underline{2.6} & 2.1
        & \cellcolor{red!7}1.7
        & \cellcolor{red!7}1.4
        & \cellcolor{red!7}1.1
        \\

        \hspace{4mm} w/o. gate
        & 2.4 & 1.4 & 2.8
        & 0.0 & 0.0 & 0.0
        & 0.0 & 0.0 & 0.0
        & 0.3 & 0.0 & 0.0
        & 1.5 & 0.1 & 0.4
        & \cellcolor{red!7}0.8
        & \cellcolor{red!7}0.3
        & \cellcolor{red!7}0.6
        \\

        \hspace{4mm} w. mean
        & 8.6 & \underline{3.4} & \textbf{3.2}
        & \underline{0.2} & 0.0 & 0.0
        & 0.0 & 0.0 & 0.0
        & 3.3 & 0.7 & 0.2
        & 5.9 & 1.3 & \underline{2.2}
        & \cellcolor{red!7}3.6
        & \cellcolor{red!7}1.1
        & \cellcolor{red!7}1.1
        \\

        \midrule

        \multicolumn{19}{l}{
            \textit{1.3B params with 100B training tokens}
        }
        \\

        \rowcolor{gray!30}
        \hspace{2mm} Transformer
        & 0.0 & 0.0 & 0.0
        & 0.0 & \underline{0.0} & \underline{0.0}
        & \underline{0.0} & \underline{0.0} & \underline{0.0}
        & 0.0 & 0.0 & 0.0
        & 0.0 & 0.0 & 0.0
        & \cellcolor{red!7}0.0
        & \cellcolor{red!7}0.0
        & \cellcolor{red!7}0.0
        \\

        \hspace{2mm} MoBA
        & 0.0 & 0.0 & 0.0
        & 0.0 & \underline{0.0} & \underline{0.0}
        & \underline{0.0} & \underline{0.0} & \underline{0.0}
        & 0.0 & 0.0 & 0.0
        & 0.0 & 0.0 & 0.0
        & \cellcolor{red!7}0.0
        & \cellcolor{red!7}0.0
        & \cellcolor{red!7}0.0
        \\

        \hspace{2mm} GSA
        & 9.4 & \underline{5.8} & 3.8
        & 0.0 & \underline{0.0} & \underline{0.0}
        & \underline{0.0} & \underline{0.0} & \underline{0.0}
        & 6.3 & \underline{4.1} & \textbf{3.1}
        & 8.3 & 5.2 & 2.5
        & \cellcolor{red!7}4.8
        & \cellcolor{red!7}\underline{3.0}
        & \cellcolor{red!7}1.9
        \\

        \hspace{2mm} KDA
        & \underline{13.0} & 2.6 & 3.2
        & \underline{3.4} & \textbf{0.2} & \underline{0.0}
        & \textbf{0.4} & \underline{0.0} & \underline{0.0}
        & \underline{9.4} & 2.4 & 1.7
        & \underline{11.0} & 3.4 & 2.4
        & \cellcolor{red!7}\underline{7.4}
        & \cellcolor{red!7}1.7
        & \cellcolor{red!7}1.5
        \\

        \hspace{2mm} NHA
        & 10.0 & 4.6 & \underline{5.8}
        & 0.0 & \underline{0.0} & \underline{0.0}
        & \underline{0.0} & \underline{0.0} & \underline{0.0}
        & 8.9 & \textbf{4.6} & 2.7
        & 9.4 & \underline{5.4} & \underline{3.0}
        & \cellcolor{red!7}5.7
        & \cellcolor{red!7}2.9
        & \cellcolor{red!7}\underline{2.3}
        \\

        \rowcolor{blue!10}
        \hspace{2mm} PHBA
        & \textbf{13.6} & \textbf{8.2} & \textbf{6.2}
        & \textbf{20.0} & \underline{0.0} & \underline{0.0}
        & \underline{0.0} & \underline{0.0} & \underline{0.0}
        & \textbf{11.5} & 2.1 & \underline{3.0}
        & \textbf{11.9} & \textbf{5.8} & \textbf{4.7}
        & \cellcolor{red!7}\textbf{11.4}
        & \cellcolor{red!7}\textbf{3.2}
        & \cellcolor{red!7}\textbf{2.8}
        \\

        \bottomrule

    \end{tabular}
    \end{adjustbox}

    % ============================================================
    % PART III: LongBench
    % ============================================================
    \vspace{0.05em}

    \renewcommand{\arraystretch}{0.98}
    \setlength{\tabcolsep}{3.0pt}

    \begin{adjustbox}{width=1.0\textwidth,center}
    \begin{tabular}{
        @{}
        l|
        cc|
        cc|
        ccc|
        cc|
        cc|
        c
        @{}
    }

        \toprule

        \multicolumn{13}{c}{
            \textbf{PART III: LongBench Tasks}
        }
        \\

        \midrule

        \multirow{2}{*}{\textbf{Model \& Scale}}
        & \multicolumn{2}{c|}{\textbf{Single-Doc}}
        & \multicolumn{2}{c|}{\textbf{Multi-Doc}}
        & \multicolumn{3}{c|}{\textbf{Summarization}}
        & \multicolumn{2}{c|}{\textbf{Few-shot}}
        & \multicolumn{2}{c|}{\textbf{Code}}
        & \multirow{2}{*}{\textbf{Avg}}
        \\

        \cmidrule(lr){2-3}
        \cmidrule(lr){4-5}
        \cmidrule(lr){6-8}
        \cmidrule(lr){9-10}
        \cmidrule(lr){11-12}

        &
        \textbf{MF.}
        & \textbf{Qasp.}
        & \textbf{Hotp.}
        & \textbf{2Wiki.}
        & \textbf{Gov.}
        & \textbf{Multi.}
        & \textbf{SAM.}
        & \textbf{TREC}
        & \textbf{Triv.}
        & \textbf{LCC}
        & \textbf{Repo.}
        &

        \\

        \midrule

        \multicolumn{13}{l}{
            \textit{760M params with 50B training tokens}
        }
        \\

        \rowcolor{blue!10}
        \hspace{2mm} PHBA (origin)
        & 14.82
        & \textbf{6.54}
        & 6.03
        & \underline{8.84}
        & \underline{8.77}
        & 8.74
        & 19.06
        & 38.67
        & 24.19
        & 12.95
        & 12.90
        & \cellcolor{red!7}14.68
        \\

        \hspace{4mm} w. $C=512, K=2$
        & 14.02
        & 4.87
        & 6.08
        & 8.61
        & 6.82
        & 6.43
        & 18.75
        & 32.67
        & 24.11
        & 15.73
        & 17.35
        & \cellcolor{red!7}14.13
        \\

        \hspace{4mm} w. $C=256, K=4$
        & 13.96
        & 5.37
        & 6.17
        & 7.86
        & 6.26
        & 8.41
        & 22.10
        & 29.33
        & 21.08
        & \textbf{16.29}
        & 18.11
        & \cellcolor{red!7}14.09
        \\

        \hspace{4mm} w. $C=64, K=16$
        & \underline{15.35}
        & \underline{6.31}
        & \textbf{7.17}
        & 8.61
        & 8.22
        & 6.97
        & \textbf{24.62}
        & \textbf{43.67}
        & \textbf{24.33}
        & 15.13
        & 14.91
        & \cellcolor{red!7}\textbf{15.94}
        \\

        \hspace{4mm} w. $M=64$
        & \textbf{16.50}
        & 5.93
        & 6.05
        & \textbf{9.33}
        & \textbf{10.85}
        & \textbf{10.55}
        & 16.80
        & 37.83
        & 17.83
        & 15.78
        & \underline{18.54}
        & \cellcolor{red!7}\underline{15.09}
        \\

        \hspace{4mm} w. $M=256$
        & 13.99
        & 5.32
        & 5.90
        & 8.78
        & 8.17
        & 8.75
        & 17.27
        & \underline{40.67}
        & 21.71
        & 14.84
        & 15.17
        & \cellcolor{red!7}14.60
        \\

        \hspace{4mm} w/o. gate
        & 12.04
        & 5.62
        & 5.91
        & \textbf{9.33}
        & 7.75
        & \underline{9.64}
        & 12.12
        & 24.00
        & 23.71
        & \underline{16.05}
        & 14.30
        & \cellcolor{red!7}12.77
        \\

        \hspace{4mm} w. mean
        & 13.71
        & 6.18
        & 5.70
        & 7.70
        & 7.49
        & 7.22
        & 19.56
        & 30.33
        & 22.85
        & 15.85
        & \textbf{19.68}
        & \cellcolor{red!7}14.21
        \\

        \midrule

        \multicolumn{13}{l}{
            \textit{1.3B params with 100B training tokens}
        }
        \\

        \rowcolor{gray!30}
        \hspace{2mm} Transformer
        & 10.97
        & 5.83
        & 4.21
        & 5.68
        & 7.60
        & 6.74
        & 10.38
        & 23.00
        & 16.15
        & 8.38
        & 8.91
        & \cellcolor{red!7}9.80
        \\

        \hspace{2mm} MoBA
        & 10.01
        & 5.80
        & 3.74
        & 6.34
        & \underline{8.63}
        & 7.43
        & 11.47
        & 25.00
        & 16.39
        & 12.07
        & 9.53
        & \cellcolor{red!7}10.58
        \\

        \hspace{2mm} GSA
        & 13.48
        & 5.33
        & 6.43
        & \underline{8.57}
        & 6.31
        & 7.37
        & 24.58
        & 19.00
        & 21.78
        & 16.49
        & 19.34
        & \cellcolor{red!7}13.52
        \\

        \hspace{2mm} KDA
        & \underline{13.64}
        & \underline{6.26}
        & \textbf{7.57}
        & 8.33
        & 7.58
        & 8.77
        & \textbf{27.20}
        & \underline{33.00}
        & \underline{22.17}
        & \underline{18.64}
        & \underline{21.43}
        & \cellcolor{red!7}\underline{15.87}
        \\

        \hspace{2mm} NHA
        & 13.63
        & 6.08
        & 5.41
        & 7.95
        & 5.96
        & \textbf{10.06}
        & 24.77
        & 32.50
        & 13.34
        & \textbf{21.54}
        & \textbf{22.78}
        & \cellcolor{red!7}14.91
        \\

        \rowcolor{blue!10}
        \hspace{2mm} PHBA
        & \textbf{15.79}
        & \textbf{7.16}
        & \underline{7.49}
        & \textbf{9.85}
        & \textbf{12.10}
        & \underline{9.26}
        & \underline{25.06}
        & \textbf{46.83}
        & \textbf{26.75}
        & 18.39
        & 20.81
        & \cellcolor{red!7}\textbf{18.14}
        \\

        \bottomrule

    \end{tabular}
    \end{adjustbox}

    \end{small}

    \label{tab:reordered_long_context_final}

\end{table*}
\subsection{Additional Experimental Details}
\label{sec:appendix-exp-details}

\subsubsection{Model Architecture and Training.}
\label{sec:m-a-t}
We train all models from scratch on FineWeb-Edu with a context length of 8K.
The 760M and 1.3B models are trained on 50B and 100B tokens, respectively,
with hidden dimensions of 1536 and 2048 and a head dimension of 128.
The 1.3B models use 16 attention heads and an MLP expansion ratio of 4.
Unless otherwise specified, PHBA uses $M=128$ prefix-state slots and
block size $C=128$. For the 1.3B main comparison, the exact-token budget is
fixed to $1/8$ of the sequence. At the 8K training length, this corresponds
to 1024 exact tokens per query, or $K=8$ blocks consisting of the current
causal block and seven routed historical blocks.

We use a 32K SentencePiece tokenizer and optimize all models with AdamW,
using $\beta_1=0.9$, $\beta_2=0.95$, a peak learning rate of
$3\times10^{-4}$, weight decay of $0.1$, and gradient clipping at $1.0$.
The learning rate follows cosine decay with a 1\% warmup.
Training uses a global batch size of 64 sequences
(524,288 tokens per optimization step) in bfloat16 on 8 NVIDIA B200-192GB GPUs.

\subsubsection{Baseline Configuration.}
\label{sec:b-c}
We compare PHBA with Transformer~\citep{vaswani2017attention},
MoBA~\citep{lu2026moba}, GSA~\citep{zhang2024gated},
KDA~\citep{team2025kimi}, and NHA~\citep{du2026native}.
To isolate the effect of the sequence-mixing mechanism, all compared
models use standalone architectures rather than hybrid layer compositions.
All baselines are trained on the same data and context length at matched
model scale, while following the recommended architectural configurations
of the corresponding methods. Sparse-attention methods use matched
exact-token budgets where applicable.

\subsubsection{Evaluation Protocol.}
\label{sec:e-p}
For short-context evaluation, we report language modeling on WikiText and
LAMBADA~\citep{merity2016pointer,paperno2016lambada}, together with zero-shot
commonsense reasoning on ARC-Easy/Challenge, HellaSwag, PIQA, and WinoGrande
~\citep{clark2018think,zellers2019hellaswag,bisk2020piqa,sakaguchi2021winogrande}.
All short-context evaluations use
\texttt{lm-evaluation-harness}~\citep{eval-harness}.

For long-context evaluation, models trained only at 8K are directly evaluated
at 16K, 32K, and 64K without further training or fine-tuning.
We use RULER~\citep{hsieh2024ruler} to evaluate retrieval and reasoning over
long sequences, including common-word extraction (CWE), question answering
based on HotpotQA and SQuAD, and controlled NIAH variants.
We further evaluate real-world long-context understanding on
LongBench~\citep{bai2024longbench}, covering document QA, summarization,
few-shot learning, and code-related tasks.
LongBench inputs of at most 8K tokens are evaluated in full; longer inputs are
truncated by retaining the first 4K and last 4K tokens.

\subsubsection{Efficiency Benchmark Protocol.}
\label{sec:e-b-p}

Due to hardware availability, the efficiency benchmarks are conducted on a
single NVIDIA H20-96GB GPU rather than the NVIDIA B200-192GB GPUs used for model
training. All compared methods are evaluated on the same H20-96GB hardware and
under the same BF16 training configuration. We benchmark end-to-end training
throughput at sequence lengths of 2K, 4K, 8K, and 16K, and report throughput
in thousands of processed tokens per second (Kt/s). For sparse-attention methods, we compare configurations under a matched
$1/8$ exact-token budget. For MoBA and PHBA, we consider two complementary
parameterizations. In the fixed-$K$ setting, $K=4$ and the block size $C$
is adjusted with sequence length to preserve the $1/8$ token budget. In the
fixed-block setting, $C=256$ and the number of selected blocks $K$ is adjusted
accordingly. This keeps the amount of exact token-level attention comparable
across sequence lengths while exposing the different routing and block-size
trade-offs. For SWA and NHA, we report both a fixed-window configuration with window size
64 and a matched-budget configuration in which the window size grows with
sequence length to maintain the same $1/8$ token budget. All state-based
methods use $M=128$ memory slots. Transformer uses full causal attention. Each measurement covers the complete training computation, including forward
and backward propagation and parameter updates. Input preparation and
data-loading overhead are excluded from the timing so that the comparison
reflects model and operator efficiency. Measurements are taken after warm-up
iterations to remove initialization effects. NHA with the $1/8$ configuration exceeds the available GPU memory from 4K
onward under the evaluated implementation; these cases are reported as OOM
(zero throughput in Fig.~\ref{fig:throughput_phba}).
\end{document}

%% file: math_commands.tex
\usepackage{amsmath,amsfonts,bm}

\def\eqref#1{equation~\ref{#1}}
\def\1{\bm{1}}

\DeclareMathAlphabet{\mathsfit}{\encodingdefault}{\sfdefault}{m}{sl}
\SetMathAlphabet{\mathsfit}{bold}{\encodingdefault}{\sfdefault}{bx}{n}